\pdfoutput=1

\documentclass[11pt,a4paper]{article}
\usepackage[dvipsnames,table]{xcolor}
\usepackage{acl} 

\usepackage{times}
\usepackage{latexsym}
\usepackage[normalem]{ulem}
\useunder{\uline}{\ul}{}
\usepackage{booktabs}
\usepackage{supertabular}
\usepackage{float}
\usepackage{graphicx}
\usepackage{enumitem}

\usepackage{microtype}

\def\aclpaperid{53} 

\definecolor{cadmiumgreen}{rgb}{0.0, 0.42, 0.24}

\newcommand{\para}[1]{\smallskip\noindent\textbf{#1}}

\title{{D}econstructing {N}LG {E}valuation: 
\\ {E}valuation {P}ractices, {A}ssumptions, and {T}heir {I}mplications}

\author{Kaitlyn Zhou \\
  Stanford University \\[-0.2em]
  \texttt{\small katezhou@stanford.edu}
  \And
  Su Lin Blodgett \\
  Microsoft Research \\[-0.2em]
  \texttt{\small sublodge@microsoft.com}
  \And
  Adam Trischler \\
  Microsoft Research \\[-0.2em]
  \texttt{\small adtrisch@microsoft.com}
  \AND
  Hal Daumé III \\
  Microsoft Research \& \\ University of Maryland \\[-0.2em]
  \texttt{\small me@hal3.name}
  \And 
  Kaheer Suleman \\
  Microsoft Research \\[-0.2em]
  \texttt{\small kasulema@microsoft.com}
  \And
  Alexandra Olteanu \\
  Microsoft Research \\[-0.2em]
  \texttt{\small aloltea@microsoft.com}
  }

\date{}

\newcommand{\statacad}[2]{\textcolor{black}{(#1\% academic, #2\% non-academic)}}

\begin{document}
\maketitle
\begin{abstract}
There are many ways to express similar things in text, which makes evaluating natural language generation (NLG) systems difficult. 
Compounding this difficulty is the need to assess varying quality criteria depending on the deployment setting.
While the landscape of NLG evaluation has been well-mapped, practitioners' goals, assumptions, and constraints---which inform decisions about what, when, and how to evaluate---are often partially or implicitly stated, or not stated at all.
Combining a formative semi-structured interview study of NLG practitioners (N=18) with a survey study of a broader sample of practitioners (N=61), 
we surface goals, community practices, assumptions, and constraints that shape NLG evaluations, examining their implications and how they embody ethical considerations.

\end{abstract}
\section{Introduction}

Evaluating natural language generation (NLG) models and systems---that generate new or altered text---is notoriously difficult, as often there are many valid ways to express similar content in text. 
This difficulty is often compounded by the need for NLG systems to meet a variety of goals---often measured in competing and imperfect ways---depending on their deployment settings~\citep{sai-etal-2021-perturbation}.
Such challenges have been mapped out for a variety of NLG tasks, systems, and settings \citep{novikova-etal-2017-need,howcroft-etal-2020-twenty,liang2021towards}, including examinations of how their design, deployment, and evaluation can give rise to ethical concerns~\citep{cercas-curry-rieser-2018-metoo,schlesinger2018let,hovy-etal-2020-sound,sheng-etal-2021-societal,abercrombie-etal-2021-alexa}.

These challenges also mean that practitioners seeking to evaluate their NLG systems must decide 
on their \textit{goals} for the system (e.g., help users write fluently), what \textit{quality criteria} to use to reflect those goals (e.g., output fluency), and how to \textit{operationalize} the measurement of those criteria (e.g., perplexity or crowd judgments). 
Such decisions are often guided by deployment settings,
by evaluation norms in a research community~\cite{gkatzia-mahamood-2015-snapshot}, 
by assumptions about ``proper'' NLG model or system behavior, as well as 
by real-world constraints~\cite{sai2020survey, gehrmann-etal-2021-gem}. 
Although the landscape of quality criteria and operationalizations has been surveyed,
much less is known about how practitioners' goals, assumptions, and constraints shape their decisions about which criteria to use, when, how, and why. 
Limited visibility into how these factors shape NLG evaluations can, however, make it difficult to anticipate issues these factors might give rise to, or even what we can learn about research progress.

In this paper, we surface goals, assumptions, community practices, and constraints that NLG practitioners\footnote{In this study, we dub ``practitioners'' all those with experience in researching, developing, deploying, or evaluating NLG models/systems. This includes both academic and industry researchers, as well as applied scientists and engineers.} work with when evaluating NLG systems, including limitations and ethical implications. 
These tacit elements of NLG evaluation are challenging to ascertain from literature surveys, as they are often unstated or only partially stated.

To reveal otherwise latent aspects of NLG evaluation practice, 
we apply a mixed-methods approach from human-computer interaction (HCI).
Specifically, we conduct a formative semi-structured interview study (N=18) and a survey study (N=61) of NLG practitioners (\S\ref{sec:methods}) in order to identify common themes related to practitioners' goals (\S\ref{sec:goals}), practices (\S\ref{subsec:practices}), and assumptions and constraints (\S\ref{subsec:rationales}) when evaluating NLG systems, including ethical considerations (\S\ref{sec:ethics}). 
By recognizing more tacit elements of NLG evaluation, our work aims to provide scaffolding for discerning the issues they may give rise to and help re-think NLG evaluations.

\section{Background and Methods}
\label{sec:methods}

To examine NLG evaluation practices---along with the goals, assumptions, and constraints that inform them---and whether they incorporate assessments of possible adverse impacts, we first conducted a formative semi-structured interview study of NLG practitioners (\S\ref{sec:interview}). 
To probe the themes emerging from our interviews at greater scale, we subsequently conducted a survey study with a broader sample of practitioners (\S\ref{sec:survey}).

\para{Participant recruitment and IRB. } We recruited NLG practitioners using snowball sampling~\cite{parker2019snowball}, via targeted emails and social media posts, with a request to share with relevant groups. Each interviewee received a $\$25$ gift card for a 45 minute video-conference interview, while survey participants could enter a raffle for one of ten $\$50$ gift cards. Both studies were IRB approved, and we obtained informed consent from all participants.

\subsection{Interview Study}
\label{sec:interview}

To design our study, we drew on the measurement modeling framework~\cite{2021-jacobs-wallach} to tease apart practitioners' conceptualization of quality criteria from how they measure those criteria.
We also investigated how current NLG evaluation practice grapples with possible adverse impacts,
such as fairness and inclusion issues~\cite{sheng-etal-2021-societal,weidinger2021ethical}.

We began our interviews with a) background questions on participants' NLG projects and experience to establish context. Subsequent questions broadly asked about b)~practitioners' overall project goals, c)~the quality criteria they want to assess their system on, and d) how they measure those criteria---with c) and d) following the measurement modeling framework. We also asked participants about adverse impacts of their work, and whether they measure or assess them. 

Before conducting the final interviews, we first ran pilot interviews (N=5) with practitioners from both product and research organizations to identify possible clarity issues with our questions and study protocol. 
The full interview protocol is available in Appendix \ref{section: script_appendix}.
Overall, we interviewed 18 practitioners from 12 organizations (Table~\ref{table:participants}).

\para{Identifying interview themes.}
To analyze interview transcripts, we used a bottom-up approach rooted in grounded theory~\cite{strauss1997grounded,charmaz2014constructing}, following \citet{robertson2021i}. 
We iteratively coded and thematically sorted interview excerpts by looking for relations with or among already assigned codes. 
Specifically, we first distributed the interview transcripts across all authors for open coding, with the resulting codes being then discussed by all authors to identify and agree upon clusters of common themes.

To organize codes into themes, we drew again on measurement modeling to distinguish between codes for what practitioners want to measure ({\em quality criteria}) and codes for how practitioners operationalize those measurements ({\em evaluation practices}). 
To examine which factors affect practitioners' decisions about what criteria they prioritize and operationalize, we also consider the {\em goals}, {\em assumptions} and {\em constraints} that underpin their evaluation practices, clustering codes into related themes. Finally, we identify themes on how practitioners conceptualize and embody possible {\em adverse impacts}.

\begin{table}[h]
\centering 
\scriptsize
\rowcolors{2}{gray!10}{white}
\begin{tabular}{p{4.5cm}cc}
\toprule
{\bf Sector}	& {\bf Interview} & {\bf Survey} \\
\midrule
Academia &	8 & 31 \\
Industry &	8 & 17 \\
Academia \& Industry & 2 & 10 \\
Gov./NGO & - & 3 \\
\midrule
{\bf Role}      & \\
\midrule
Professor  & 2 & 12 \\
Researcher or Postdoc & 9  & 16 \\
Graduate Student & 5 & 17 \\
Manager & 1 & 3 \\
Engineer or Data Scientist & 1 & 6 \\
Other & - & 7 \\
\midrule
\rowcolor{white}Total & 18 & 61 \\
\bottomrule
\end{tabular}
\vspace{-6pt}
\caption{Overview of interview and survey participants.}
\label{table:participants}
\vspace{-8pt}
\end{table}

\subsection{Survey Study}
\label{sec:survey}

We designed our survey around themes from the interview study, starting again with questions about participants' background, and organizing the remaining questions under a few broad themes---a)~quality criteria and goals, b) evaluation practices, c)~evaluation rationales (including assumptions and constraints), and d) adverse impacts and ethical considerations---with different sets of questions matching different themes. 
Similarly with the interview study, before sharing and advertising our survey, we piloted it with a few industry researchers (N=3) to identify possible clarity issues with our survey questions and protocol.  
See Appendix \ref{section: survey_questions_appendix} for full survey script. 
To filter out spammers, we relied on 3 of the open-text questions---the description of participants' occupation, description of their NLG project, and their reflections on how to improve NLG evaluation.
After removing spammers, we analyzed responses from 61 participants (Table~\ref{table:participants}).

\section{Interview and Survey Findings}
Across both studies, we observe common themes in participants' responses that we overview below. 
Interview participant quotes---anonymized and paraphrased for brevity and clarity---are followed by ``P'' and a participant ID, while quotes from survey responses are only marked by ``SP.'' 
Statistics based on survey responses are followed by ``SQ'' and the question number.
We report disaggregated statistics as~\statacad{X}{Y} when gaps exist between academic and non-academic participants. 
We note, however, that the statistics we report are only indicative as our respondent sample is skewed towards industry and academic researchers (Table~\ref{table:participants}), and not necessarily representative of the broader NLG community.




\subsection{Evaluation Goals and Quality Criteria}
\label{sec:goals}
What practitioners want their NLG systems or models to do ({\em goals}) and what they aim to assess to determine whether those goals are met ({\em quality criteria}) both shape which evaluation practices they are likely to follow. 
Participants' reflections on what success looks like for their projects and which quality criteria they deem critical could thus help discern the implications of the resulting practices.

\para{The goals of NLG work are often disconnected from deployment settings or users.}
Our participants describe diverse goals, like imbuing bots with social skills or
helping users write fluently.
However, many practitioners do not have clarity about or consider possible deployment settings or users, e.g., \textit{``[W]e were not really thinking [about users], usually I try not to think about [users]''}~[P8]. 
This is especially true when participants had no deployment plans, including in non-academic settings---26\% of all survey participants \statacad{39}{13}~[SQ16].
A lack of clarity about possible deployment scenarios, however, can make it difficult to assess how appropriate certain metrics are~\cite{blodgett-etal-2021-stereotyping} or even to imagine adverse impacts~\cite{boyarskaya2020overcoming}. 

\para{While practitioners conduct NLG evaluations to report to a variety of audiences, paper reviewers and the research community are most often mentioned.}
While we see differences between academic and non-academic participants---with the latter thinking more about specific audiences like their team or manager 
\statacad{19}{53} or other teams \statacad{16}{50}---77\% of survey participants say one goal of their evaluation was to report results in a paper \statacad{90}{63} and 59\% to report to a research community \statacad{67}{50} [SQ13]. If reporting results in papers or to research communities represents a significant goal of many practitioners, however, then evaluations already legible to a research community may be over-incentivized, and quality criteria and measurements unfamiliar or uncommon to that community may be discouraged.


\para{Quality criteria like correctness, usability, coherence, and fluency are often prioritized.}
While our participants mention a diversity of criteria, 
77\% of survey respondents deemed correctness among most important criteria \statacad{83}{70}, followed by 49\% usability \statacad{38}{60}, 47\% coherence, and 44\% fluency; with criteria like readability (21\%) and clarity (21\%) being least frequently selected [SQ14]. 
Interview participants mention similar criteria, e.g., \textit{``[W]ithout [even] figuring out how we measure [these criteria], we want the [text] to be informative [and] fluent''}~[P4]. 
This corroborates prior work that found task usefulness, fluency, grammaticality, relevance, or generic notions of output quality among most commonly used criteria for NLG evaluation \cite{howcroft-etal-2020-twenty,liang2021towards}.
This may stem from an assumption that criteria like correctness, coherence, and fluency, as intrinsic properties of text, are easier to measure and better-suited to automatic metrics \cite{gkatzia-mahamood-2015-snapshot} than more extrinsic criteria related to how text is used (i.e., usability).

\para{User engagement is often foregrounded in deployment settings.}
When systems are deployed, practitioners want \textit{``users to be happy [and] engaged''}~[P12] and may prioritize feedback and engagement-based metrics,
like \textit{``whether [the generated suggestions] would be accepted by the end user''}~[SP] 
and \textit{``the number [of] clicks [or] of times people accept your suggestions''}~[P7]. 
However, users and their feedback may not be representative and may, e.g., \textit{``skew[] heavily towards enterprise customers''}~[P12]. 
In and of itself, focusing on specific, non-representative groups is not necessarily problematic, but designing to meet the needs of an existing user base might have exclusionary outcomes if that user base is not reflective of a larger population.

\para{Practitioners may conflate different quality criteria if they lack a shared understanding of their meanings, or if the criteria lack clear conceptualizations.} 
As in prior work \cite{belz-etal-2020-disentangling,gkatzia-mahamood-2015-snapshot,howcroft-etal-2020-twenty,liang2021towards,sai2020survey}, we see a tendency to conflate different quality criteria. 
A participant noted that \textit{``[linguists tell] me [fluency and grammaticality] are clearly distinct from each other [but in NLP] most people don't know that distinction''}~[P8], and another that \textit{``a lot of these descriptive [terms for quality criteria] have overlap''} which is particularly noticeable when \textit{``trying to evaluate them by hand''}~[P3].
Different understandings of criteria or the conflation of criteria may have consequences beyond the concern that good measurements require clarity about what is being measured~\cite{blodgett-etal-2021-stereotyping}.
If e.g., grammaticality is understood as adherence to Mainstream U.S. English grammar rules (often the case due to standard language ideologies \cite{rosa2017unsettling}), then it will exclude other language varieties, like Appalachian English \cite{luhman1990appalachian} or African American Language \cite{lanehart2015oxford} and thus their speakers. Given the focus on fluency, this exclusion may be compounded if grammaticality and fluency are conflated.



\para{Quality criteria are also conflated with measurements of those criteria.} 
When asked about key performance criteria, some participants instead talk about what metrics they use, e.g., \textit{``we have to make [a text] generator that's good [and] we ensured the perplexity [was] on par with the state of the art perplexity. Perplexity was [the] performance metric we [used]''}~[P17]. 
To determine if perplexity is a good metric for the ``goodness'' of a text generator, however, we must clearly articulate what quality criteria a good generator should meet. Without this clarity, we may focus too strongly on a metric---like perplexity---and risk losing the ability to assess the metric's appropriateness and distinguish between the metric and the original goal. 

\subsection{Evaluation Practices} 
\label{subsec:practices}
Existing evaluation practices can also provide insight into 
what is deemed important to evaluate, what is overlooked during evaluation, and how quality criteria are being operationalized.

\para{Although routine, the use of automatic metrics may lack justification and validation,} as \textit{``[e]veryone is in this in-between phase of just using everything they can find''}~[P3]. 
Automatic metrics commonly include training/validation loss, perplexity, and string or word matching metrics (e.g., BLEU, ROUGE, METEOR, NIST), with 85\% of survey participants reporting always or often using automatic metrics \statacad{94}{77}~[SQ18]. 
Metrics developed for one task (\textit{``BLEU was [developed for] MT''}~[P3]) are also often adapted or used as-is for other tasks (\textit{``we [compute] BLEU for language modeling''}~[P8]), 
with 57\% of participants saying they always or often use metrics developed for other NLG tasks \statacad{68}{47}~[SQ18]. This practice can be risky if metrics validated in one setting do not capture key criteria in other settings.


\para{Different types of human evaluation are often used to complement automatic metrics;} one participant noted they conduct \textit{``human evaluation[s] just 'cause NLG metrics are still so messy''}~[P3].
When our interview participants discussed what human evaluations they conducted, they described several types, including crowdsourcing, user feedback or engagement metrics, expert evaluations, and qualitative examinations by practitioners or their colleagues. Overall, 89\% of survey participants report always or often using \textit{some} form of human evaluation \statacad{94}{83}~[SQ18]. This also means ``human evaluation'' is often imprecisely used to refer to many types of evaluations and settings.

\para{Automatic metrics often gate which models are evaluated by crowdsourcing.}
While 34\% of survey participants report always or often performing crowdsourced evaluations~[SQ18], many interview participants also say they select models based on
their performance on automatic metrics before investing in crowdsourced evaluations, since \textit{``[there's a] huge search space [and we] pick those that get good automatic metric values''}~[P12]. 
Indeed, 47\% of those who report performing crowdsourced evaluations also note crowdsourcing is always or often contingent on performing well on automatic metrics \statacad{57}{36}~[SQ18].
However, this practice might overlook models for which human evaluations anti-correlate with automatic metrics, as \textit{``automatic metrics may show something and the human evaluation may show the complete opposite. Unless you do human evaluations, you will not have a thorough picture of [how well your system is doing]''}~[P2].

\para{Human evaluations may draw on usage or feedback data in deployed settings, but many practitioners lack access to either deployment settings or usage data.}
30\% of survey participants report conducting evaluations with users in deployed settings, with a sizeable gap among academics (12\%) and non-academics (46\%)~[SQ18].
Interview participants also noted \textit{``this gap between academic [evaluations and the] things that matter when you actually deploy something out in the world''}~[P15]. 
Since feedback from users of a deployed system might more accurately reflect their in-situ needs than feedback from e.g., crowdworkers, practitioners without access to deployment settings may miss critical evaluation data and have little visibility into possible adverse impacts.

\para{Practitioners manually assess generated texts, but rarely follow a formal method when doing so.}
For some participants, manual assessment is more of an art than a science, and many do not follow any established methods or a self-devised protocol when doing so: \textit{``it just comes down to me reading a lot of samples and then choosing the one which overall seems to be better''}~[P5]. 
Indeed, while 77\% of survey participants report always or often manually examining generated texts, of those who perform such qualitative examinations, only 38\% report always or often applying a formal method~\statacad{30}{47}~[SQ18], 
and even fewer describe such a method, with a few using a labeling scheme or a sampling protocol for selecting which examples to assess [SQ19].
While common, the practice of (informally and) manually examining generated text to assess quality is rarely detailed in the literature.
It can be problematic if results of error analyses~\cite{van2021underreporting}
are shown without describing how assessments were made. 
Practitioners also may not be representative of those who use NLG systems or are reflected in generated text~\cite{hovy2016social}.

\para{While expert evaluations are sought after, it can be unclear what type of expertise is needed.}
Practitioners also seek out evaluators with certain expertise, like English teachers or linguists, e.g., \textit{``we hire English teachers [who are] experts, do a much better job and give very detailed feedback''}~[P8].
In our survey, 25\% of participants report seeking out expert judges for evaluation purposes \statacad{32}{17}~[SQ18].
Yet even ``experts'' may need to \textit{``[see] a lot of [generated] text and [internalize] the sort of a language [a model tends to produce]''}~[P17]. 

\para{Practitioners often know little about those represented in or contributing to data creation and annotation, or to crowdsourced evaluation.}
Only 18\% of survey participants report always or often having demographic information about the annotators or authors who created the datasets they use [SQ18].
Of those who report using crowdsourcing, only 30\% report always or often having demographic information about crowdsourcing judges [SQ18].
Such information can provide crucial insight into the speakers, communities, language varieties, and perspectives represented in and shaping datasets; without it, some datasets---whose text may read to some as if \textit{``it's written by [a] small group of people''}~[P17]---risk being used as representative while excluding wide swaths of people.


\para{Many practitioners re-purpose datasets created for other tasks,} with 59\% of survey participants reporting always or often doing so [SQ18], while only 48\% of survey participants report always or often using datasets created in-house for their project \statacad{42}{53}~[SQ18]. 
Datasets' suitability should however be carefully interrogated before being re-purposed. If datasets custom-built or validated for other tasks do not properly capture the new task's peculiarities or the desired quality criteria, they can pose threats to the validity of the task and its evaluation. 

\subsection{Rationales for Evaluation Practices}
\label{subsec:rationales}

To further probe participants about assumptions, constraints and other considerations that shape their evaluation practices, we also explicitly asked them about what guided their NLG evaluations.  

\para{While many practitioners recognize their limitations, automatic metrics are often assumed reliable enough. }
A participant noted they \textit{``wouldn't trust any sort of automatic measure of a text generation system [as they need] more than just a good BLEU or ROUGE score before [they'd] sign off on using a language model''}~[P11], 
while others questioned whether automatic metrics \textit{``capture anything meaningful''}~[P13] when assessing latent constructs like creativity.

Despite these and other documented shortcomings \cite{gkatzia-mahamood-2015-snapshot,novikova-etal-2017-need,kuribayashi-etal-2021-lower,liang2021towards}, practitioners do rely broadly on automatic metrics: 50\% of survey participants agree or strongly agree that automatic metrics represent reliable ways to assess NLG systems or models~[SQ20], while 
43\% say that metrics developed for one NLG task can be reliably used or adapted to evaluate other NLG tasks \statacad{32}{53}~[SQ22]. 
One participant remarked that \textit{``automatic metric[s are] still more scalable and objective than human evaluation''}~[SP]. 
This trust in automatic metrics' reliability across tasks may entrench their use, possibly crowding out other measurement approaches, especially as publication processes may reward metrics that allow for easy comparison to existing models. 

\para{When it comes to automatic metrics, many practitioners believe more is better. } A participant noted \textit{``the current status is you kind of just throw it all at the wall. So if there's another automatic metric, great, let's add that to the suite of metrics we're running''}~[P13], and another that they \textit{``ended up having a lot of metrics because none of them is a superset of any others. They all have issues''}~[P17]. 
54\% of survey participants also agree or strongly agree it is a good practice to evaluate with as many automatic metrics as possible~[SQ22]. 
Yet a ``kitchen sink'' approach to evaluation risks obfuscating quality with quantity, yielding apparent model improvements that may not correlate with actual performance, or may result in expectations to see improvements on certain metrics, even when those may not be necessary or desirable.
This approach may exist partly to pay a ``tech debt,'' where metrics are included mainly to compare to legacy baselines assessed with them.

\para{Human evaluations are often assumed to be the gold standard, but the term refers to many things.}
A participant said they \textit{``always trust the human evaluations, that's the gold standard. If the human evaluation is not great, then we would not consider that a good solution''}~[P9]. 
This was also echoed by survey participants, with 69\% believing automatic metrics should always or often be judged by how they correlate with human evaluations \statacad{74}{63}~[SQ23].
Most participants assume human evaluations can be reliably used for NLG evaluation, with 56\% agreeing or strongly agreeing that is the case for crowdsourcing; 71\% for manual inspection by practitioners themselves; and 89\% for metrics based on usage patterns in deployment settings~[SQ20]. 
Given these variations in practitioners' trust in different types of human evaluations, the blanket use of the term ``human evaluation'' to refer to them collectively can make it hard to calibrate claims about human evaluation results.

\para{Expert evaluations are believed to help elicit insights that are hard to obtain via crowdsourcing.}
Without proper training, a layperson might struggle to adequately assess generated text \cite{clark-etal-2021-thats, karpinska-etal-2021-perils}. One participant noted people \textit{``often get very impressed by neural text when researchers might be aware that it looks impressive because it stay[s] safe [and] just say[s] generic things''}~[P17]. 
In our survey, 87\% of participants believe experts always or often contribute to the evaluation in ways standard crowdsourcing cannot [SQ23]. 
Experts may however be even less representative of those using a NLG system and may inject their shared biases into evaluations \cite{van2019best}.


\para{Practitioners say NLG evaluations are most constrained by datasets, annotators, lack of metrics, and inability to deploy.} 
When asked to rank resources by how much they constrain their evaluations, 59\% of survey participants ranked relevant datasets among their top 3 most limiting resources \statacad{48}{70}, 50\% ranked expert annotators, 42\% the ability to deploy, 37\% crowdsourcing costs \statacad{45}{30}, 
36\% relevant metrics; with compute and engineering resources being least frequently ranked among top limiting resources. 

\para{Existing practices, metrics and data shape the evaluation choices of many practitioners.} 
Participants noted that \textit{``academic benchmarks are very important [as] they're [an] easy measure to figure out whether your experiments are working''}~[P15] or that \textit{``we used all the basic standard automatic [metrics in] that suit''}~[P2].
When asked to rank considerations that guide their evaluations, 
55\% of survey participants ranked among their top 3 existing standards and practices \statacad{67}{43}, 36\% ranked dataset and metrics availability \statacad{45}{26}, and 29\% maintaining performance in deployed settings \statacad{12}{46}~[SQ26]. 
This emphasis on existing practices, metrics and data may be due to challenges to developing them, expectations of audiences like a research community or a team, or to make it easy to compare with prior work.
This reliance also makes their proper development and validation even more critical.

\para{Poor performance is most often attributed to poor data quality,} with some practitioners also believing \textit{``one [way] of driving quality upwards is by feeding in more data to these models''}~[P12]. Others echoed this belief, explaining how their \textit{``model tends to talk about things similar to its training data; whatever you feed into the model, that's what the model [will] output''}~[P7] as \textit{``language models are kind of bottom up, you have a pile of historical data and you learn the patterns of language used in that data''}~[P11]. 
In our survey, 62\% of practitioners agree or strongly agree that output quality is primarily impacted by the quality of the training data \statacad{55}{70}~[SQ22]. 
Practitioners may thus rely mostly on expanding or curating datasets to improve performance, overlooking other approaches (e.g., optimization methods, UX design) or offloading issues onto dataset builders~\cite{sambasivan2021everyone}.

\para{Practitioners believe datasets should be formally evaluated, but that rarely happens.}
71\% of survey participants say there is always or often a need to evaluate the quality of training datasets \statacad{77}{63}, 
and 87\% say so for evaluation datasets \statacad{97}{77}~[SQ23].
An interview participant, however, pointed to the \textit{``concern of who is in charge of validating all these datasets. I'm sure some slightly harmful things can fall through the cracks, that no one takes a look at just due to the sheer size of these datasets''}~[P3]. 
Datasets that are not carefully evaluated may not be well-matched to what they aim to measure \cite{blodgett-etal-2021-stereotyping} or give rise to harms \cite[e.g.,][]{birhane2021large}.
\subsection{Ethical Considerations and Impacts}
\label{sec:ethics}
Finally, we solicited practitioners' reflections on ethical issues or adverse impacts stemming from their projects, and if they try to measure or mitigate such issues. We use the term \textit{fairness and inclusion} (F\&I) to refer to a broad range of adverse impacts, like allocational and representational harms~\cite{barocas2017problem,blodgett-etal-2020-language}.
Instead of imposing our definitions, we deliberately asked broadly about  F\&I issues and other adverse impacts and ethical considerations to understand how practitioners conceptualize them. 



\para{Practitioners worry about adverse impacts, but NLG evaluations rarely include assessments of F\&I issues.} 
A participant emphasized that \textit{``developing [an] ethically correct [system that] takes care of things like safety and even fairness and inclusion is really important, [but] tackling that is not trivial''}~[P10]. 
While this interest in possible adverse impacts is encouraging, about 37\% of survey participants thought their project had no adverse impacts~[SQ28], and only 20\% report systematically measuring F\&I issues beyond toxicity \statacad{10}{30}~[SQ31].
Furthermore, while some practitioners were \textit{``glad things like [stereotyping benchmarks] are coming out, [although those are] pretty insufficient,''}~[P12] and \textit{``there are currently no good evaluation practices for [F\&I issues]''}~[SP].
Thus, their \textit{``metrics don't capture [F\&I] right now''}~[P1] or were not \textit{``specifically designed to capture [F\&I] issues''}~[P8].

\para{To incorporate F\&I in their evaluations, practitioners want more ready-to-use and scalable metrics.} 
Participants also described how there are \textit{``20 metrics trying to measure gender and occupation [bias], but that's so narrow and [it's only] one indicator we have [and] often we don't know how [it] might relate to other forms of bias''}~[P15],  
and how there is no \textit{``ImageNet[-like option] for responsible AI [to do things at that] scale for language''}~[P7] that \textit{``practitioners and product teams [can] leverage in an easier way to scale these things up''}~[P14].
A notable 34\% of survey participants also agree or strongly agree it is possible to develop automatic metrics to adequately assess F\&I~[SQ33]. 
This preference for scale and ease-of-use, however, may nudge practitioners towards relying on new metrics before their validity and reliability are properly assessed~\cite[e.g.,][]{blodgett-etal-2021-stereotyping}.

\para{For some practitioners, F\&I is a concern only for deployed models or systems.} A participant noted they did not think \textit{``much [about ethical issues, as] the models [they] built never got deployed anywhere, never will most likely [get deployed], [they were] mostly just a proof of concept''}~[P8].
Furthermore, 34\% of survey respondents agree or strongly agree F\&I issues are less important for academic prototypes \statacad{26}{43}, while 18\% agree or strongly agree F\&I is only relevant if NLG systems are deployed, and 33\% that F\&I issues can only be effectively measured in deployed settings \statacad{26}{40}~[SQ33].
Models and prototypes are however sometimes repurposed, and deployed systems built upon prototypes or academic work not meant to be deployed may give rise to F\&I issues.

\para{F\&I work is often seen as separate from NLG work, or deferred to someone else.}
36\% of survey participants agree or strongly agree F\&I is a separate research area from NLG evaluation \statacad{26}{47}~[SQ33].
Interview participants also echoed this, with even those doing F\&I work \textit{``having [separate papers] dedicated to studying toxicity or bias in language''} since \textit{``the only concerns reviewers [have] is how many benchmarks we ran our model [on], they're not concerned about whether our [model] is more or less biased than [other models]''}~[P17]. Participants also defer the responsibility of F\&I work to others working, e.g., on user facing features and \textit{``who [will later] further refine [a model's output],''} or to researchers as \textit{``product team[s don't] get enough time to research [F\&I] metrics''}~[P1]. 
Deferring F\&I work to later stages of NLG development or to others may produce workflows where teams must frequently react and correct arising issues.






\para{Ethical considerations are often reduced to ``toxic'' speech.} 
When probed about ethical issues, many interview participants were concerned about generating or reproducing hateful, offensive, or toxic language: \textit{``a big consideration is [that] a lot of the data [could] in some ways be toxic content of different varieties''}~[P15] and \textit{``so often what these models will do is generate something that is very very unsafe and in some cases toxic''}~[P10]. 
This narrow focus on toxicity may inadvertently crowd out other F\&I issues.
Surprisingly, despite this focus on toxic language, only 30\% of survey participants reported systematically evaluating whether generated text contains such language \statacad{23}{37}~[SQ31], meaning even these issues might be under-reported.



\para{Practitioners hold a range of beliefs about what properties the language in their datasets or generated by NLG systems should embody.}
52\% of survey participants agree or strongly agree that generated text should adhere to standard grammar rules of the language it is generated in~[SQ33], and 57\% say neutrality is a property NLG systems should embody~[SQ37]. 
Many also {\em want} their system or model to capture either a ``neutral voice'' (30\%), ``no voice'' (8\%), or not to capture any particular writing or speaking voice (33\% overall; 45\% academic, 20\% non-academic)~[SQ35].

By contrast, only 18\% believed the datasets they use do not capture any voice,
and 18\% say they capture a neutral voice~[SQ34], while 21\% also believe their systems or models do not capture any voice, and 16\% say they capture a neutral voice \statacad{10}{23}~[SQ36]. 
Notably, while 13\% of non-academic participants think their system captures a specific, non-neutral voice, none of the academic participants do; this is also reflected in the interviews, where multiple industry participants aim to capture a specific type of voice like matching a \textit{``company voice''}~[P6].
%
Such beliefs about language 
are consistent with pervasive language ideologies---``the cultural system of ideas about social and linguistic relationships'' 
\cite{irvine1989when}---including that some kinds of language (and their speakers) are inherently correct, neutral, or appropriate for public use~\cite{rosa2017language}, and thus that generated language can be neutral or voiceless. 
Since assumptions about language are rarely named or interrogated, it remains unclear how they shape practitioners' evaluations.
But they can give rise to harm; for example, a belief that race-related language is inherently inappropriate if generated \cite{schlesinger2018let} can erase language by or topics important to minoritized users.


%


%


\para{The topics and linguistic style of generated text are rarely systematically measured.}
When asked if they measured the linguistic style of generated text, a participant, under the assumption that \textit{``these models [just] follow the training data,''} noted they \textit{``don't necessarily analyze the generated output''}~[P9].
Only 25\% of survey participants report systematically measuring what topics their system tends to generate, while 18\% report systematically measuring the linguistic style of generated text \statacad{10}{27}~[SQ31] (notably, though, 43\% report informally assessing style or tone, e.g., via eyeballing).
Coupled with assumptions about neutral or voiceless language, this suggests that
the over- or under-representation of topics, perspectives, or language varieties may not be generally conceptualized as F\&I issues
and measured. 
NLG systems may, however, unintentionally prioritize ``voices'' over-represented in datasets,
especially when assumed neutral and not examined.
\para{To mitigate F\&I issues, practitioners seek control over what is generated,}
as many do not want \textit{``an uncontrolled model to spew out random information''}~[P2]. Others want their systems \textit{``to divert the conversation into [less sensitive] domains''}~[P6] by \textit{``trying to control [what is generated using] different [e.g.] types of sampling''}~[P7]. 
While such mechanisms may prevent the generation of some problematic texts, most are just \textit{``blunt means of controlling the generation''}~[P12], and can lead to e.g., erasure harms if topics important to minoritized groups are deemed inappropriate~\cite{schlesinger2018let,dodge-etal-2021-documenting}: \textit{``[R]efusing to talk about these issues probably means you are distributionally refusing to talk about things marginalized populations care about more''}~[P5]. 
\para{Blocklists are often the go-to mitigation technique for F\&I issues.}
A majority of interview participants report using blocklists as an F\&I mitigation approach---including to control system outputs---and 21\% of survey participants report always or often using blocklists \statacad{7}{37}~[SQ30], with notably more use in non-academic settings.
However, as one participant noted, \textit{``we have blocklists and phrasal blocklists [to help] prevent some of the major categories of failure, and they’re understandable and explainable and adjustable, which is the upside. But is the coverage good? I don't think it is''}~[P12]. 
\section{Conclusion}
We identify practices and assumptions that shape the evaluations of NLG systems, which we argue deserve further attention due to their potential adverse impacts, particularly when those practices and assumptions go partly stated, or are not stated at all. 
Our findings suggest a number of approaches and future directions, including some raised by survey participants when asked how to improve NLG evaluation~[SQ38].


First, the range of implicit assumptions and choices we uncover indicates that future evaluations might be strengthened by making these assumptions and choices explicit, including how quality criteria are conceptualized, how manual analysis is conducted and shapes decisions, and whether some evaluation approaches gate others (as automatic evaluations often gate human ones). 

Participants also suggested several opportunities for evaluation; one respondent raised the possibility of lab studies to explore people's perceptions and uncover failure modes without fully deploying a system, while another emphasized the need to study systems in interaction given the importance of social and task context. Industry participants described a range of practices, including user engagement metrics, red  teaming, and blocklists, which may be less well-studied in academia.

The challenges surrounding the use of automatic metrics---particularly the pressure towards a ``kitchen sink'' approach despite widespread recognition of many existing metrics' limitations---may not have an easy remedy. 
A ``kitchen sink'' approach should be thoroughly justified, and not used only because practitioners lack clarity about their goals and quality criteria. 
Beyond clarity about goals and quality criteria, a better understanding of evaluation datasets and metrics---whose perspectives and language are contained in datasets, what datasets and metrics can and cannot capture, and how metrics are connected to user or downstream outcomes---may also clarify their appropriateness for different settings, and disincentivize simply evaluating with as many of them as possible.



Furthermore, human evaluations, particularly those grounded in specific deployment settings or involving extensive participation, tend to be costly. 
Not only is such work expensive, but it may also involve 
qualitative or participatory approaches less familiar to NLP practitioners (than using automatic metrics), or may slow down product releases or publications. 
Industry teams often constrained by tight deadlines may be especially unlikely to adopt qualitative or participatory approaches in the absence of readily available, well-validated methods and guidance for their use. 
While such approaches will never be as inexpensive as many automatic metrics, user studies or design workshops with users to help develop and solidify methods and guidance might lower barriers to their broader adoption. Finally, shifting publication incentives towards work grounded in deployment scenarios, as well as work engaging with methods from other disciplines (e.g., HCI or sociolinguistics) may encourage the investment of time and resources needed to carry out more  thorough evaluations.





Turning to ethical considerations, our participants often raised the need for investment in the development of resources for conceptualizing and measuring F\&I issues. 
These include better frameworks of harms (to help anticipate issues beyond toxicity), 
appropriate datasets, 
additional frameworks to guide (and scale) qualitative evaluations, 
task-specific metrics to combat automatic metric re-use without proper validation, 
and approaches for measuring latent qualities such as voice, style, and topic (which remain longstanding open questions).
Other opportunities include broader disciplinary shifts. 
Several participants desired clearer community standards addressing not only best practices in modeling, 
but also meta-level questions---e.g., who makes decisions on what ethical considerations to prioritize (for evaluation and beyond), or how to develop repeatable mechanisms for surfacing, reporting, and addressing failures as they arise.


\section{Ethical Considerations}

Our interview and survey studies---which are intended to uncover community assumptions, constraints, and practices---necessarily surface those of some but not all practitioners. Although we aimed to recruit as widely as possible, our participants were recruited via snowball sampling, seeded with researchers and industry groups that were well-known to us, as well as on social media (and therefore biased by our followers). 
Additionally, our interviews and surveys were only in English.
Those who participated largely focus on research, work on English language NLG systems, and live and work in the Global North. While this may reflect the NLG research community and who has been able to participate in it,
our findings may not reflect the assumptions or practices of practitioners working in other settings.

The results of our studies are also rather descriptive, and (though we think they are comprehensive) they cannot provide an exhaustive set of assumptions, practices, or F\&I concerns, and should be interpreted accordingly. While our goal is to encourage further reflections about practices, the factors that shape them, and possible implications, we may risk instead inadvertently discouraging certain evaluation practices.

\section*{Acknowledgements}
We sincerely thank Adam Atkinson, Alessandro Sordoni, Angelina Wang, Chad Atalla, Dallas Card, Eric Yuan, Jackie Chi Kit Cheung, Lucy Li, Marion Zepf, Tong Wang, 
the MSR FATE Group, and our anonymous reviewers for their support, and helpful reviews and feedback. We also want to thank our interview and survey participants for taking the time to participate in our research and for sharing their experiences with us. This research has been supported in part by a Stanford Graduate Fellowship.

\bibliographystyle{acl_natbib}
\bibliography{acl2021}

\begin{thebibliography}{37}
\expandafter\ifx\csname natexlab\endcsname\relax\def\natexlab#1{#1}\fi

\bibitem[{Abercrombie et~al.(2021)Abercrombie, Cercas~Curry, Pandya, and
  Rieser}]{abercrombie-etal-2021-alexa}
Gavin Abercrombie, Amanda Cercas~Curry, Mugdha Pandya, and Verena Rieser. 2021.
\newblock \href {https://doi.org/10.18653/v1/2021.gebnlp-1.4} {{A}lexa,
  {G}oogle, {S}iri: What are your pronouns? gender and anthropomorphism in the
  design and perception of conversational assistants}.
\newblock In \emph{Proceedings of the 3rd Workshop on Gender Bias in Natural
  Language Processing}, pages 24--33, Online. Association for Computational
  Linguistics.

\bibitem[{Barocas et~al.(2017)Barocas, Crawford, Shapiro, and
  Wallach}]{barocas2017problem}
Solon Barocas, Kate Crawford, Aaron Shapiro, and Hanna Wallach. 2017.
\newblock \href
  {http://meetings.sigcis.org/uploads/6/3/6/8/6368912/program.pdf} {The problem
  with bias: Allocative versus representational harms in machine learning}.
\newblock \emph{In Proceedings of SIGCIS, Philadelphia, PA.}

\bibitem[{Belz et~al.(2020)Belz, Mille, and
  Howcroft}]{belz-etal-2020-disentangling}
Anya Belz, Simon Mille, and David~M. Howcroft. 2020.
\newblock \href {https://aclanthology.org/2020.inlg-1.24} {Disentangling the
  properties of human evaluation methods: A classification system to support
  comparability, meta-evaluation and reproducibility testing}.
\newblock In \emph{Proceedings of the 13th International Conference on Natural
  Language Generation}, pages 183--194, Dublin, Ireland. Association for
  Computational Linguistics.

\bibitem[{Blodgett et~al.(2020)Blodgett, Barocas, Daum{\'e}~III, and
  Wallach}]{blodgett-etal-2020-language}
Su~Lin Blodgett, Solon Barocas, Hal Daum{\'e}~III, and Hanna Wallach. 2020.
\newblock \href {https://doi.org/10.18653/v1/2020.acl-main.485} {Language
  (technology) is power: A critical survey of {``}bias{''} in {NLP}}.
\newblock In \emph{Proceedings of the 58th Annual Meeting of the Association
  for Computational Linguistics}, pages 5454--5476, Online. Association for
  Computational Linguistics.

\bibitem[{Blodgett et~al.(2021)Blodgett, Lopez, Olteanu, Sim, and
  Wallach}]{blodgett-etal-2021-stereotyping}
Su~Lin Blodgett, Gilsinia Lopez, Alexandra Olteanu, Robert Sim, and Hanna
  Wallach. 2021.
\newblock \href {https://doi.org/10.18653/v1/2021.acl-long.81} {Stereotyping
  {N}orwegian salmon: An inventory of pitfalls in fairness benchmark datasets}.
\newblock In \emph{Proceedings of the 59th Annual Meeting of the Association
  for Computational Linguistics and the 11th International Joint Conference on
  Natural Language Processing (Volume 1: Long Papers)}, pages 1004--1015,
  Online. Association for Computational Linguistics.

\bibitem[{Boyarskaya et~al.(2020)Boyarskaya, Olteanu, and
  Crawford}]{boyarskaya2020overcoming}
Margarita Boyarskaya, Alexandra Olteanu, and Kate Crawford. 2020.
\newblock \href {https://arxiv.org/pdf/2011.13416.pdf} {Overcoming failures of
  imagination in {AI} infused system development and deployment}.
\newblock \emph{Navigating the Broader Impacts of AI Research Workshop at
  NeurIPS 2020}.

\bibitem[{Cercas~Curry and Rieser(2018)}]{cercas-curry-rieser-2018-metoo}
Amanda Cercas~Curry and Verena Rieser. 2018.
\newblock \href {https://doi.org/10.18653/v1/W18-0802} {{\#}{M}e{T}oo {A}lexa:
  How conversational systems respond to sexual harassment}.
\newblock In \emph{Proceedings of the Second {ACL} Workshop on Ethics in
  Natural Language Processing}, pages 7--14, New Orleans, Louisiana, USA.
  Association for Computational Linguistics.

\bibitem[{Charmaz(2014)}]{charmaz2014constructing}
Kathy Charmaz. 2014.
\newblock \emph{Constructing grounded theory}.
\newblock SAGE.

\bibitem[{Clark et~al.(2021)Clark, August, Serrano, Haduong, Gururangan, and
  Smith}]{clark-etal-2021-thats}
Elizabeth Clark, Tal August, Sofia Serrano, Nikita Haduong, Suchin Gururangan,
  and Noah~A. Smith. 2021.
\newblock \href {https://doi.org/10.18653/v1/2021.acl-long.565} {All that{'}s
  {`}human{'} is not gold: Evaluating human evaluation of generated text}.
\newblock In \emph{Proceedings of the 59th Annual Meeting of the Association
  for Computational Linguistics and the 11th International Joint Conference on
  Natural Language Processing (Volume 1: Long Papers)}, pages 7282--7296,
  Online. Association for Computational Linguistics.

\bibitem[{Dodge et~al.(2021)Dodge, Sap, Marasovi{\'c}, Agnew, Ilharco,
  Groeneveld, Mitchell, and Gardner}]{dodge-etal-2021-documenting}
Jesse Dodge, Maarten Sap, Ana Marasovi{\'c}, William Agnew, Gabriel Ilharco,
  Dirk Groeneveld, Margaret Mitchell, and Matt Gardner. 2021.
\newblock \href {https://doi.org/10.18653/v1/2021.emnlp-main.98} {Documenting
  large webtext corpora: A case study on the colossal clean crawled corpus}.
\newblock In \emph{Proceedings of the 2021 Conference on Empirical Methods in
  Natural Language Processing}, pages 1286--1305, Online and Punta Cana,
  Dominican Republic. Association for Computational Linguistics.

\bibitem[{Gehrmann et~al.(2021)Gehrmann, Adewumi, Aggarwal, Ammanamanchi,
  Aremu, Bosselut, Chandu, Clinciu, Das, Dhole, Du, Durmus, Du{\v{s}}ek,
  Emezue, Gangal, Garbacea, Hashimoto, Hou, Jernite, Jhamtani, Ji, Jolly, Kale,
  Kumar, Ladhak, Madaan, Maddela, Mahajan, Mahamood, Majumder, Martins,
  McMillan-Major, Mille, van Miltenburg, Nadeem, Narayan, Nikolaev,
  Niyongabo~Rubungo, Osei, Parikh, Perez-Beltrachini, Rao, Raunak, Rodriguez,
  Santhanam, Sedoc, Sellam, Shaikh, Shimorina, Sobrevilla~Cabezudo, Strobelt,
  Subramani, Xu, Yang, Yerukola, and Zhou}]{gehrmann-etal-2021-gem}
Sebastian Gehrmann, Tosin Adewumi, Karmanya Aggarwal, Pawan~Sasanka
  Ammanamanchi, Anuoluwapo Aremu, Antoine Bosselut, Khyathi~Raghavi Chandu,
  Miruna-Adriana Clinciu, Dipanjan Das, Kaustubh Dhole, Wanyu Du, Esin Durmus,
  Ond{\v{r}}ej Du{\v{s}}ek, Chris~Chinenye Emezue, Varun Gangal, Cristina
  Garbacea, Tatsunori Hashimoto, Yufang Hou, Yacine Jernite, Harsh Jhamtani,
  Yangfeng Ji, Shailza Jolly, Mihir Kale, Dhruv Kumar, Faisal Ladhak, Aman
  Madaan, Mounica Maddela, Khyati Mahajan, Saad Mahamood, Bodhisattwa~Prasad
  Majumder, Pedro~Henrique Martins, Angelina McMillan-Major, Simon Mille, Emiel
  van Miltenburg, Moin Nadeem, Shashi Narayan, Vitaly Nikolaev, Andre
  Niyongabo~Rubungo, Salomey Osei, Ankur Parikh, Laura Perez-Beltrachini,
  Niranjan~Ramesh Rao, Vikas Raunak, Juan~Diego Rodriguez, Sashank Santhanam,
  Jo{\~a}o Sedoc, Thibault Sellam, Samira Shaikh, Anastasia Shimorina,
  Marco~Antonio Sobrevilla~Cabezudo, Hendrik Strobelt, Nishant Subramani, Wei
  Xu, Diyi Yang, Akhila Yerukola, and Jiawei Zhou. 2021.
\newblock \href {https://doi.org/10.18653/v1/2021.gem-1.10} {The {GEM}
  benchmark: Natural language generation, its evaluation and metrics}.
\newblock In \emph{Proceedings of the 1st Workshop on Natural Language
  Generation, Evaluation, and Metrics (GEM 2021)}, pages 96--120, Online.
  Association for Computational Linguistics.

\bibitem[{Gkatzia and Mahamood(2015)}]{gkatzia-mahamood-2015-snapshot}
Dimitra Gkatzia and Saad Mahamood. 2015.
\newblock \href {https://doi.org/10.18653/v1/W15-4708} {A snapshot of {NLG}
  evaluation practices 2005 - 2014}.
\newblock In \emph{Proceedings of the 15th {E}uropean Workshop on Natural
  Language Generation ({ENLG})}, pages 57--60, Brighton, UK. Association for
  Computational Linguistics.

\bibitem[{Hovy et~al.(2020)Hovy, Bianchi, and
  Fornaciari}]{hovy-etal-2020-sound}
Dirk Hovy, Federico Bianchi, and Tommaso Fornaciari. 2020.
\newblock \href {https://doi.org/10.18653/v1/2020.acl-main.154} {{``}you sound
  just like your father{''} commercial machine translation systems include
  stylistic biases}.
\newblock In \emph{Proceedings of the 58th Annual Meeting of the Association
  for Computational Linguistics}, pages 1686--1690, Online. Association for
  Computational Linguistics.

\bibitem[{Hovy and Spruit(2016)}]{hovy2016social}
Dirk Hovy and Shannon~L. Spruit. 2016.
\newblock \href {https://doi.org/10.18653/v1/P16-2096} {The social impact of
  natural language processing}.
\newblock In \emph{Proceedings of the 54th Annual Meeting of the Association
  for Computational Linguistics (Volume 2: Short Papers)}, pages 591--598,
  Berlin, Germany. Association for Computational Linguistics.

\bibitem[{Howcroft et~al.(2020)Howcroft, Belz, Clinciu, Gkatzia, Hasan,
  Mahamood, Mille, van Miltenburg, Santhanam, and
  Rieser}]{howcroft-etal-2020-twenty}
David~M. Howcroft, Anya Belz, Miruna-Adriana Clinciu, Dimitra Gkatzia, Sadid~A.
  Hasan, Saad Mahamood, Simon Mille, Emiel van Miltenburg, Sashank Santhanam,
  and Verena Rieser. 2020.
\newblock \href {https://aclanthology.org/2020.inlg-1.23} {Twenty years of
  confusion in human evaluation: {NLG} needs evaluation sheets and standardised
  definitions}.
\newblock In \emph{Proceedings of the 13th International Conference on Natural
  Language Generation}, pages 169--182, Dublin, Ireland. Association for
  Computational Linguistics.

\bibitem[{Irvine(1989)}]{irvine1989when}
Judith~T. Irvine. 1989.
\newblock \href {http://www.jstor.org/stable/645001} {When talk isn't cheap:
  Language and political economy}.
\newblock \emph{American Ethnologist}, 16(2):248--267.

\bibitem[{Jacobs and Wallach(2021)}]{2021-jacobs-wallach}
Abigail~Z. Jacobs and Hanna Wallach. 2021.
\newblock \href {https://doi.org/10.1145/3442188.3445901} {Measurement and
  fairness}.
\newblock \emph{Proceedings of the 2021 ACM Conference on Fairness,
  Accountability, and Transparency}.

\bibitem[{Karpinska et~al.(2021)Karpinska, Akoury, and
  Iyyer}]{karpinska-etal-2021-perils}
Marzena Karpinska, Nader Akoury, and Mohit Iyyer. 2021.
\newblock \href {https://doi.org/10.18653/v1/2021.emnlp-main.97} {The perils of
  using {M}echanical {T}urk to evaluate open-ended text generation}.
\newblock In \emph{Proceedings of the 2021 Conference on Empirical Methods in
  Natural Language Processing}, pages 1265--1285, Online and Punta Cana,
  Dominican Republic. Association for Computational Linguistics.

\bibitem[{Kuribayashi et~al.(2021)Kuribayashi, Oseki, Ito, Yoshida, Asahara,
  and Inui}]{kuribayashi-etal-2021-lower}
Tatsuki Kuribayashi, Yohei Oseki, Takumi Ito, Ryo Yoshida, Masayuki Asahara,
  and Kentaro Inui. 2021.
\newblock \href {https://doi.org/10.18653/v1/2021.acl-long.405} {Lower
  perplexity is not always human-like}.
\newblock In \emph{Proceedings of the 59th Annual Meeting of the Association
  for Computational Linguistics and the 11th International Joint Conference on
  Natural Language Processing (Volume 1: Long Papers)}, pages 5203--5217,
  Online. Association for Computational Linguistics.

\bibitem[{Lanehart(2015)}]{lanehart2015oxford}
Sonja Lanehart. 2015.
\newblock \emph{The Oxford Handbook of African American Language}.
\newblock Oxford University Press.

\bibitem[{van~der Lee et~al.(2019)van~der Lee, Gatt, van Miltenburg, Wubben,
  and Krahmer}]{van2019best}
Chris van~der Lee, Albert Gatt, Emiel van Miltenburg, Sander Wubben, and Emiel
  Krahmer. 2019.
\newblock \href {https://doi.org/10.18653/v1/W19-8643} {Best practices for the
  human evaluation of automatically generated text}.
\newblock In \emph{Proceedings of the 12th International Conference on Natural
  Language Generation}, pages 355--368, Tokyo, Japan. Association for
  Computational Linguistics.

\bibitem[{Liang and Li(2021)}]{liang2021towards}
Hongru Liang and Huaqing Li. 2021.
\newblock Towards standard criteria for human evaluation of chatbots: A survey.
\newblock \emph{arXiv preprint arXiv:2105.11197}.

\bibitem[{Luhman(1990)}]{luhman1990appalachian}
Reid Luhman. 1990.
\newblock Appalachian english stereotypes: language attitudes in kentucky.
\newblock \emph{Language in Society}, 19(3):331--348.

\bibitem[{Novikova et~al.(2017)Novikova, Du{\v{s}}ek, Cercas~Curry, and
  Rieser}]{novikova-etal-2017-need}
Jekaterina Novikova, Ond{\v{r}}ej Du{\v{s}}ek, Amanda Cercas~Curry, and Verena
  Rieser. 2017.
\newblock \href {https://doi.org/10.18653/v1/D17-1238} {Why we need new
  evaluation metrics for {NLG}}.
\newblock In \emph{Proceedings of the 2017 Conference on Empirical Methods in
  Natural Language Processing}, pages 2241--2252, Copenhagen, Denmark.
  Association for Computational Linguistics.

\bibitem[{Parker et~al.(2019)Parker, Scott, and Geddes}]{parker2019snowball}
Charlie Parker, Sam Scott, and Alistair Geddes. 2019.
\newblock \href
  {https://doi.org/https://dx.doi.org/10.4135/9781526421036831710} {Snowball
  sampling}.
\newblock \emph{SAGE research methods foundations}.

\bibitem[{Prabhu and Birhane(2021)}]{birhane2021large}
Vinay~Uday Prabhu and Abeba Birhane. 2021.
\newblock \href
  {https://ieeexplore.ieee.org/stamp/stamp.jsp?tp=&arnumber=9423393&tag=1}
  {Large image datasets: A pyrrhic win for computer vision?}
\newblock \emph{2021 IEEE Winter Conference on Applications of Computer Vision
  (WACV)}, pages 1536--1546.

\bibitem[{Robertson et~al.(2021)Robertson, Olteanu, Diaz, Shokouhi, and
  Bailey}]{robertson2021i}
Ronald Robertson, Alexandra Olteanu, Fernando Diaz, Milad Shokouhi, and Peter
  Bailey. 2021.
\newblock \href
  {https://www.microsoft.com/en-us/research/publication/i-cant-reply-with-that-characterizing-problematic-email-reply-suggestions/}
  {"{I} can’t reply with that": Characterizing problematic email reply
  suggestions}.
\newblock In \emph{CHI Conference on Human Factors in Computing Systems (CHI
  ’21)}. ACM.

\bibitem[{Rosa and Burdick(2017)}]{rosa2017language}
Jonathan Rosa and Christa Burdick. 2017.
\newblock \href {https://doi.org/10.1093/oxfordhb/9780190212896.013.15}
  {\emph{{Language Ideologies}}}, pages 103--124. Oxford University Press.

\bibitem[{Rosa and Flores(2017)}]{rosa2017unsettling}
Jonathan Rosa and Nelson Flores. 2017.
\newblock \href {https://doi.org/10.1017/S0047404517000562} {Unsettling race
  and language: Toward a raciolinguistic perspective}.
\newblock \emph{Language in Society}, 46(5):621–647.

\bibitem[{Sai et~al.(2021)Sai, Dixit, Sheth, Mohan, and
  Khapra}]{sai-etal-2021-perturbation}
Ananya~B. Sai, Tanay Dixit, Dev~Yashpal Sheth, Sreyas Mohan, and Mitesh~M.
  Khapra. 2021.
\newblock \href {https://doi.org/10.18653/v1/2021.emnlp-main.575} {Perturbation
  {C}heck{L}ists for evaluating {NLG} evaluation metrics}.
\newblock In \emph{Proceedings of the 2021 Conference on Empirical Methods in
  Natural Language Processing}, pages 7219--7234, Online and Punta Cana,
  Dominican Republic. Association for Computational Linguistics.

\bibitem[{Sai et~al.(2022)Sai, Mohankumar, and Khapra}]{sai2020survey}
Ananya~B. Sai, Akash~Kumar Mohankumar, and Mitesh~M. Khapra. 2022.
\newblock \href {https://doi.org/10.1145/3485766} {A {S}urvey of {E}valuation
  {M}etrics {U}sed for {NLG} {S}ystems}.
\newblock \emph{ACM Comput. Surv.}, 55(2).

\bibitem[{Sambasivan et~al.(2021)Sambasivan, Kapania, Highfill, Akrong,
  Paritosh, and Aroyo}]{sambasivan2021everyone}
Nithya Sambasivan, Shivani Kapania, Hannah Highfill, Diana Akrong, Praveen
  Paritosh, and Lora~M Aroyo. 2021.
\newblock \href {https://doi.org/10.1145/3411764.3445518} {“everyone wants to
  do the model work, not the data work”: Data cascades in high-stakes ai}.
\newblock In \emph{Proceedings of the 2021 CHI Conference on Human Factors in
  Computing Systems}, CHI '21, New York, NY, USA. Association for Computing
  Machinery.

\bibitem[{Schlesinger et~al.(2018)Schlesinger, O'Hara, and
  Taylor}]{schlesinger2018let}
Ari Schlesinger, Kenton~P. O'Hara, and Alex~S. Taylor. 2018.
\newblock \href {https://doi.org/10.1145/3173574.3173889} {Let's talk about
  race: Identity, chatbots, and ai}.
\newblock In \emph{Proceedings of the 2018 CHI Conference on Human Factors in
  Computing Systems}, CHI '18, page 1–14, New York, NY, USA. Association for
  Computing Machinery.

\bibitem[{Sheng et~al.(2021)Sheng, Chang, Natarajan, and
  Peng}]{sheng-etal-2021-societal}
Emily Sheng, Kai-Wei Chang, Prem Natarajan, and Nanyun Peng. 2021.
\newblock \href {https://doi.org/10.18653/v1/2021.acl-long.330} {Societal
  biases in language generation: Progress and challenges}.
\newblock In \emph{Proceedings of the 59th Annual Meeting of the Association
  for Computational Linguistics and the 11th International Joint Conference on
  Natural Language Processing (Volume 1: Long Papers)}, pages 4275--4293,
  Online. Association for Computational Linguistics.

\bibitem[{Strauss and Corbin(1997)}]{strauss1997grounded}
Anselm Strauss and Juliet~M Corbin. 1997.
\newblock \emph{Grounded theory in practice}.
\newblock Sage.

\bibitem[{Van~Miltenburg et~al.(2021)Van~Miltenburg, Clinciu, Du{\v{s}}ek,
  Gkatzia, Inglis, Lepp{\"a}nen, Mahamood, Manning, Schoch, Thomson
  et~al.}]{van2021underreporting}
Emiel Van~Miltenburg, Miruna Clinciu, Ond{\v{r}}ej Du{\v{s}}ek, Dimitra
  Gkatzia, Stephanie Inglis, Leo Lepp{\"a}nen, Saad Mahamood, Emma Manning,
  Stephanie Schoch, Craig Thomson, et~al. 2021.
\newblock Underreporting of errors in nlg output, and what to do about it.
\newblock In \emph{Proceedings of the 14th International Conference on Natural
  Language Generation}, pages 140--153.

\bibitem[{Weidinger et~al.(2021)Weidinger, Mellor, Rauh, Griffin, Uesato,
  Huang, Cheng, Glaese, Balle, Kasirzadeh et~al.}]{weidinger2021ethical}
Laura Weidinger, John Mellor, Maribeth Rauh, Conor Griffin, Jonathan Uesato,
  Po-Sen Huang, Myra Cheng, Mia Glaese, Borja Balle, Atoosa Kasirzadeh, et~al.
  2021.
\newblock Ethical and social risks of harm from language models.
\newblock \emph{arXiv preprint arXiv:2112.04359}.

\end{thebibliography}

\appendix
\newpage
\appendix

\onecolumn
\begin{center}
    \Large \textbf{Appendix for:\\Deconstructing NLG Evaluation:\\Evaluation Practices, Assumptions, and Their Implications}
\end{center}

\section{Interview Script}
\label{section: script_appendix}
\subsection{Introduction} 
We are conducting a semi-structured interview regarding your experience with natural language generation models and systems evaluation. We hope to learn about general practices, goals, constraints and assumptions researchers and practitioners make when evaluating NLG systems. Before we start, could I get verbal consent from you that it is okay to record this meeting and take notes?  

We aim to have the interviews last ~45 minutes, and we will cover about 15 high-level questions. Finally, we would love to hear your thoughts about this process and what can be improved. We will also send you a gift card via email.  
Do you have any questions before we begin?

\subsection{Background Information}

[IQ1] Could you describe your experience with developing, deploying, and researching NLG systems?

\noindent
[IQ2] Of the projects you’ve been involved with, can you describe one or two of them?

\begin{itemize}[noitemsep,topsep=0pt,parsep=0pt,partopsep=0pt]
\item What does it do? Why is it needed?  
\item Who are your customers or stakeholders? 
\item Try to focus on one, but feel free to bring in examples and anecdotes from others as you see fit. 
\end{itemize}

\subsection{What are the stakeholders and goals?} 

[IQ3] What performance criteria are important for your system/project/model? 

\begin{itemize}[noitemsep,topsep=0pt,parsep=0pt,partopsep=0pt]
    \item What does success look like for your system/project/model? 
    \item What are the goals of your system/project/model? 
    \item Who are you optimizing for? 
\end{itemize}

\noindent
[IQ4] Can you describe the decision-making process for evaluating NLG models/systems on your team?
\begin{itemize}[noitemsep,topsep=0pt,parsep=0pt,partopsep=0pt]
    \item For example, what does the process of adding a new evaluation metric look like? 
    \item  What constraints do you have on your project? 
    \item  Can you describe the makeup of your team? What is your role? 
\end{itemize}

\noindent
[IQ5] What are you trying to demonstrate with these results?
\begin{itemize}[noitemsep,topsep=0pt,parsep=0pt,partopsep=0pt]
    \item Who are you trying to convince?
\end{itemize}

\noindent
[IQ6] Are there any ethical issues or adverse impacts that you are concerned about in the context of your project?
\begin{itemize}[noitemsep,topsep=0pt,parsep=0pt,partopsep=0pt]
    \item Are there also any F\&I concerns you have related to your system/project/model? 
\end{itemize}

\subsection{What do you want to measure?}   
[IQ7] Which criteria does your evaluation aim to capture? Possible criteria can include: fluency, readability, coherence, naturalness, quality, correctness, usability, clarity, informativeness, accuracy. Feel free to add others.


\noindent
[IQ8] [If nothing on fairness] Are your metrics trying to capture issues related to ethical AI or NLP? This can include qualities or metrics related to or for fairness and inclusion.
\begin{itemize}[noitemsep,topsep=0pt,parsep=0pt,partopsep=0pt]
    \item Fairness issues: e.g., do you look for performance disparities, bias, lack of representation 
    \item Inclusion issues: e.g., politeness, professionalism, accessibility, hate speech 
    \item What are barriers to evaluating for F\&I? 
    \item What F\&I issues are and aren't being captured well? 
\end{itemize}

\noindent
[IQ9] 
How would you characterize your experience with practices and metrics that aim at examining issues related to ethical AI/NLP? Possible answers could be: no experience, a bit of experience, I have worked on it, the focus of my work

\subsection{What do you measure? How?}  

[IQ10] Are you using automatic metrics like BLEU, ROUGE, METEOR, NIST in the development process? When?
\begin{itemize}[noitemsep,topsep=0pt,parsep=0pt,partopsep=0pt]
    \item Which ones? Why? 
    \item Are you using metrics based on some sort of “gold” or “reference” responses/datasets? 
    \item Are your datasets made in-house specifically for your project or re-used from prior work? 
    \item Do you have metadata about who created the references and how they were created? 
    \item Is your system optimizing for a specific sort of prototypical voice or persona that is captured by these references?
\end{itemize}

\noindent
[IQ11] Are you using human evaluations? When?
\begin{itemize}[noitemsep,topsep=0pt,parsep=0pt,partopsep=0pt]
    \item If applicable, is this contingent on a certain performance on automatic metrics? 
    \item If system is deployed, do you conduct evaluations with real users? How? 
    \item What is your typical budget for evaluations both time and money? 
    \item Which criteria do these human evaluations capture that your automatic metrics miss? 
\end{itemize}

\noindent
[IQ12] If your evaluation metrics give conflicting results, which metrics do you trust?
\begin{itemize}[noitemsep,topsep=0pt,parsep=0pt,partopsep=0pt]
    \item Which evaluation metrics do you consider to be the most useful in your development process? 
\end{itemize}

\noindent
[IQ13] Are there any other approaches you use to evaluate?

\noindent
[IQ14] Do you try to measure how representative the content produced by your NLG system/model is?
\begin{itemize}[noitemsep,topsep=0pt,parsep=0pt,partopsep=0pt]
    \item What does your model avoid “talking” about and what it tends to “talk” about? 
    \item Do you use red teaming to test for worst-case scenarios and/or blocklists to avoid known issues? If no, can you describe what it could be? 
    \item Are there any noticeable tendencies in the content produced by your system\/model? 
\end{itemize}

\noindent
[IQ15] Do you measure the linguistic style and tone of your model?

\begin{itemize}[noitemsep,topsep=0pt,parsep=0pt,partopsep=0pt]
    \item Are you aware of whether the content produced by your system/model has a certain tone? 
    \item Is there a specific linguistic style or tone that you are aiming for? 
    \item Do you measure what the style or tone of the content generated by your model is? 
    \item Does it tend to represent a certain type of voice or persona? 
    \item If applicable, do you know if users respond differently to the different voices captured by your system/model? If no, can you describe what it could be? 
\end{itemize}

\subsection{Reflections on your practices} 

[IQ16] Finally, are you aware of other evaluation practices? Are your practices in line? Is there anything else you’d like us to know?

\section{Online Survey Questions}
\label{section: survey_questions_appendix}
\subsection{Consent Form}

[SQ1] Do you understand and consent to these terms? {\bf R:} Yes, no 

\noindent
[SQ2] Please type your initial here if we may contact you in the future to request consent for uses of your identifiable data that are not covered in this consent form. {\bf R:} Write-in

\noindent
[SQ3] Please type your initial here if we may contact you in the future with information about follow-up or other future studies. {\bf R:} Write-in

\subsection{Background}

[SQ4] Do you have experience with developing, deploying, or researching natural language generation (NLG) systems or tasks?
\begin{itemize}[noitemsep,topsep=0pt,parsep=0pt,partopsep=0pt]
\item No, I have never worked with or on NLG 
\item I have limited experience with NLG 
\item Developing, deploying and researching NLG is part of my day-to-day job 
\item I am an expert in NLG
\end{itemize}

\noindent
[SQ5] Do you have experience with evaluating natural language generation systems or tasks?

\begin{itemize}[noitemsep,topsep=0pt,parsep=0pt,partopsep=0pt]
\item No, I have never worked with or on NLG evaluation 
\item I have limited experience with NLG evaluation 
\item NLG evaluation is part of my day-to-day job 
\item I am an expert in NLG evaluation
\end{itemize}

\noindent
[SQ6] What type of NLG system(s) or task(s) have you worked on or with? Check all that apply. 

\begin{itemize}[noitemsep,topsep=0pt,parsep=0pt,partopsep=0pt]
\item Machine translation, Paraphrasing and summarization, Question answering, Style transfer, Image captioning, Storytelling and narrative generation, Writing assistant, Content and text planning, Affect/emotion generation, Dialogue system, NLG for embodied agents and robots, Prompt completion, Other 
\end{itemize}

\noindent
[SQ7] What is the title of your occupation? {\bf R:} Write-in


\noindent
[SQ8] What sector(s) are you employed in? (Select all that apply) 

\begin{itemize}[noitemsep,topsep=0pt,parsep=0pt,partopsep=0pt]
\item Academia, Industry, Government, Non-profit, Other
\end{itemize}

\noindent
[SQ9] How many years of experience do you have in your current field of occupation? 

\begin{itemize}[noitemsep,topsep=0pt,parsep=0pt,partopsep=0pt]
\item Less than 1 year, 1 – 4 years, 5 – 9 years, 10-20 years, More than 20 years 
\end{itemize}

\noindent
[SQ10] Is research publication the goal of your work? 

\begin{itemize}[noitemsep,topsep=0pt,parsep=0pt,partopsep=0pt]
\item Never, Rarely, Sometimes, Often, Always 
\end{itemize}

\noindent
[SQ11] How many people are working on your particular NLG system or task? (How big is your team? If your project is open-sourced, approximately how many regular contributors do you have?) 

\begin{itemize}[noitemsep,topsep=0pt,parsep=0pt,partopsep=0pt]
\item 1-2, 3-4, 5-9, 10-19, 20+ 
\end{itemize}

\noindent
[SQ12] In a couple of sentences, what is the overarching goal of your NLG system or task? Feel free to include additional pertinent context for the project such as domain, dataset, task, user base etc. If you have worked on multiple NLG systems and/or tasks, please pick the one you have the most experience with, in terms of evaluation. Please answer the remainder of the survey questions with this system in mind. {\bf R:} Write-in


\subsection{Quality criteria and goals: what do you want and try to measure at evaluation time?} 

For the following questions, please consider the NLG system or task you briefly described earlier. 

\noindent
[SQ13] When you evaluate your system/model, what are the goals of this evaluation? Check all that apply. 

\begin{itemize}[noitemsep,topsep=0pt,parsep=0pt,partopsep=0pt]
\item Evaluation for reporting in a paper. 
\item Evaluation for reporting to a research community.
\item Evaluation for reporting to your team or manager.
\item Evaluation for reporting to other teams or third-party platforms. 
\item Evaluation for model development, tuning and hyperparameter search. 
\item Evaluation for ensuring the on-going performance of a deployed system.
\item Write-in
\end{itemize}

\noindent
[SQ14] Of the following, which are the most important performance criteria to evaluate for your NLG system or task? Select up to three or write your own. For more detailed definitions, please refer to the appendix here: https://aclanthology.org/2020.inlg-1.23.pdf 

\begin{itemize}[noitemsep,topsep=0pt,parsep=0pt,partopsep=0pt]
\item Fluency (the output text ‘flows well’, and is not a sequence of unconnected parts) 
\item Readability (the output text is easy to read) 
\item Coherence (the output text is presented in a well-structured, logical and meaningful way)  
\item Naturalness (the output text is likely to be used or be written by a native speaker)  
\item Correctness (the output text is correct/true relative to the input)  
\item Usability (the output text is usable in the context of a given task/application)  
\item Clarity (the output text is easy to understand)  
\item Informativeness (the amount of information conveyed by the output text)  
\item Write-in 
\end{itemize}

\noindent
[SQ15] Are there other criteria (than those listed above) that are important for your NLG system or task? {\bf R:} Write-in


\noindent
[SQ16] Is your work (or will your work be) part of an NLG system that is deployed to users in real settings (users who use the system for their own reasons, self-incentivized, and are not paid as test subjects/crowd workers)? 

\begin{itemize}[noitemsep,topsep=0pt,parsep=0pt,partopsep=0pt]
    \item Yes, no, maybe 
\end{itemize}

\noindent
[SQ17] Is your goal or one of your future goals to personalize the voice of your system for users? By voice we mean that it may capture a particular style of speaking or writing, or a particular persona. 

\begin{itemize}[noitemsep,topsep=0pt,parsep=0pt,partopsep=0pt]
    \item Yes, no, maybe 
\end{itemize}

\subsection{Evaluation practices: what do you do during evaluation? }
[SQ18] For the following statements, please answer how often you or others on your team do the following. 

Never, Rarely, Sometimes, Often, Always, (N/A) 

\begin{itemize}[noitemsep,topsep=0pt,parsep=0pt,partopsep=0pt]
    \item Use automatic evaluation metrics 
    \item Use automated metrics originally developed for other NLG tasks (off the shelf). 
    \item Use online crowdsourcing for model/system evaluation (e.g., Mechanical Turk) 
    \item If used, online crowdsourcing evaluation is contingent on your system or model achieving a certain performance level on some automatic metrics. 
    \item During development, you or others on your team manually examine or evaluate the outputs of your NLG system or model. 
    \item Follow a formal method or procedure when you or your team members are manually inspecting/evaluating the system/model output (beyond just “eyeballing”; e.g., applying a coding scheme) 
    \item Evaluate with users in actual deployed settings (users who use the system for their own reasons, self-incentivized, and are not paid as test subjects/crowd workers) 
    \item Seek out judges or annotators with certain specific expertise to assist with your system or model evaluation (e.g., linguistics) 
    \item Use datasets that were made in-house specifically for your project 
    \item Re-use datasets (as-is or modified) that were created for other tasks or  work (typically by others outside your team) 
    \item Have demographic information about the judges, annotators, or writers who created the datasets or reference solutions you use 
    \item Have demographic information about online crowdsourcing judges or annotators who evaluated the outputs of your system/task 
\end{itemize}

\noindent
[SQ19] If you follow a formal method or procedure for your manual evaluation or inspection of the system/model output, could you please describe your method? {\bf R:} Write-in


\subsection{Evaluation rationales: why do you follow certain evaluation practices?}

\noindent
[SQ20] For the system or task, you described earlier, do you agree or disagree that the following evaluation methods are reliable ways to assess performance?  Strongly agree, agree, neither agree nor disagree, disagree, strongly disagree 
\begin{itemize}[noitemsep,topsep=0pt,parsep=0pt,partopsep=0pt]
    \item Using standard automatic metrics (e.g., BLEU, ROUGE, METEOR, NIST, perplexity) 
    \item Researchers and practitioners themselves manually examining the system/ model output  
    \item Using online crowdsourced evaluations (e.g., Mechanical Turk) 
    \item Metrics based on how the system or model is being used in actual deployed settings (by users who use the system for their own reasons, self-incentivized, and are not are paid as test subjects/crowd workers) 
\end{itemize}

\noindent
[SQ21] For the system or task, you described earlier, other than those listed above, are there other evaluation methods you believe are (or would be) reliable ways to assess the performance of your system or on your task?  {\bf R:} Write-in

\noindent
[SQ22] In general, please answer if you agree or disagree with the following statement about evaluating NLG systems or tasks: Strongly agree, agree, neither agree or disagree, disagree, strongly disagree 

\begin{itemize}[noitemsep,topsep=0pt,parsep=0pt,partopsep=0pt]
    \item Metrics developed for one NLG task or system can be reliably used or adapted to evaluate other NLG tasks or systems. 
    \item It is a good practice to evaluate NLG systems or models on as many automatic metrics as possible.
    \item The quality of NLG systems or models' output is primarily impacted by the quality of the training data.
    \item Online crowdsourced evaluations are a good proxy for evaluations with users. 
\end{itemize}

\noindent
[SQ23] In general, please answer how often the following statements are true: Never, Rarely, Sometimes, Often, Always, (N/A) 

\begin{itemize}[noitemsep,topsep=0pt,parsep=0pt,partopsep=0pt]
    \item Automatic metrics should be judged by the degree to which they correlate with human evaluations (e.g., online crowdsourced, expert evaluation, team evaluation).  
    \item Judges or annotators with specific qualifications or expertise can contribute to the evaluation in ways that standard crowdsourcing cannot. 
    \item The quality of training datasets needs to be formally evaluated. 
    \item The quality of evaluation datasets needs to be formally evaluated. 
\end{itemize}

\subsection{Constraints and other considerations}

[SQ24] Rank which of the following resources currently limit the evaluation for your particular system or task the most. Top choice being the most limiting resource. 

\begin{itemize}[noitemsep,topsep=0pt,parsep=0pt,partopsep=0pt]
    \item Insufficient access to relevant datasets for evaluation 
    \item Insufficient access to relevant metrics for evaluation 
    \item Insufficient access to online paid crowdsourcing (due to cost, availability, time, or any other reason)  
    \item Insufficient access to being able to deploy systems or models to real world settings 
    \item Insufficient access to users who use the system for their own reasons (self-incentivized, rather than because they are paid as test subjects/crowd workers)  
    \item Insufficient access to judges or annotators with certain qualifications or expertise 
    \item Insufficient access to compute resources 
    \item Insufficient access to engineering resources (e.g., lack of engineers, lack of time) 
    \item Other 
\end{itemize}

\noindent
[SQ25] If there are other limiting resources for evaluation, please elaborate. {\bf R:} Write-in

\noindent
[SQ26] Rank which of the following considerations currently guide your evaluation practices the most. Top choice being the most influential consideration.

\begin{itemize}[noitemsep,topsep=0pt,parsep=0pt,partopsep=0pt]
    \item Potential concerns raised by research paper reviewers 
    \item Community guidance, standards, or practices 
    \item The need to convince yourself or your team 
    \item The need to convince other stakeholders 
    \item Availability of relevant datasets and metrics
    \item Financial considerations  
    \item The number of anticipated users that the system is expected to reach 
    \item Public relations/public image  
    \item The need for the system to perform well in real deployment settings (i.e., ecological validity)
    \item Other
    
\end{itemize}

\noindent
[SQ27] If there are other considerations for evaluation, please elaborate. {\bf R:} Write-in


\subsection{Adverse impacts and ethical considerations}

\noindent
[SQ28] Are there any ethical issues or adverse impacts that you're concerned about in the context of your project, and if so, what are they?  {\bf R:} Write-in


\noindent
[SQ29] Are there any of the ethical issues or adverse impacts you mentioned above that are particularly hard to evaluate or that you are unable to evaluate? If so, what are they?  {\bf R:} Write-in

\subsection{Adverse impacts and ethical considerations (continued)}

\noindent
[SQ30] Do you intentionally try to block certain type of content from your generation? (e.g., by using blocklists or classifiers, cleaning the training data, etc.) 

\begin{itemize}[noitemsep,topsep=0pt,parsep=0pt,partopsep=0pt]
    \item Never, Rarely, Sometimes, Often, Always, (N/A)
\end{itemize}

\noindent
[SQ31] For the following questions, for your system, how do you measure the following? Fairness and Inclusion states that NLG systems should treat all people equally, empowering and engaging everyone by providing equal benefit and access to e.g., opportunities and resources. Systematically, Informally (eyeballing), Indirectly (through another metric), No evaluation/not sure 

\begin{itemize}[noitemsep,topsep=0pt,parsep=0pt,partopsep=0pt]
    \item Which topics your system or model tends to generate and what it tends to avoid generating 
    \item The linguistic style or tone of the content generated by your system or model 
    \item Toxic, hateful, or offensive language produced by your system or model 
    \item Other fairness and inclusion issues aside from toxic, hateful, and offensive language 
\end{itemize}

\noindent
[SQ32] If you assess other fairness and inclusion issues, please briefly explain what do you measure and how.  {\bf R:} Write-in


\noindent
[SQ33] In general, for the following statements, please answer if you agree or disagree with each statement for NLG systems’ evaluation. Strongly agree, agree, neither agree or disagree, disagree, strongly disagree 

\begin{itemize}[noitemsep,topsep=0pt,parsep=0pt,partopsep=0pt]
    \item It’s possible to develop automatic metrics that can reliably evaluate the fairness and inclusion of a system. 
    \item Fairness and inclusion metrics are only relevant for NLG systems that are deployed to real users. 
    \item Fairness and inclusion is a separate research area from NLG evaluation.
    \item The language an NLG system or model generates should adhere to standard grammar rules of the language the content is generated in. E.g., an English model should adhere to the grammar rules of standard English. 
    \item Fairness and inclusion is less important for academic prototypes. 
    \item Fairness and inclusion of a system can only be effectively measured in a real deployed setting. 
\end{itemize}

\noindent
[SQ34] Do the training or evaluation dataset(s) you use for your system or model capture a particular voice? By voice we mean that it may capture a particular style of speaking or writing, or a particular persona.  

\begin{itemize}[noitemsep,topsep=0pt,parsep=0pt,partopsep=0pt]
    \item They capture a non-neutral, specific voice (a particular persona or style). 
    \item They capture an assemblage of voices. 
    \item They capture a neutral voice.
    \item They do not capture any voice.
    \item They could be capturing a voice but I’m not sure what it is. 
\end{itemize}

\noindent
[SQ35] Do you aim for your system or model to capture a particular voice? 

\begin{itemize}[noitemsep,topsep=0pt,parsep=0pt,partopsep=0pt]
    \item We aim for it to capture a non-neutral, specific voice (a particular persona). 
    \item We aim for it to capture an assemblage of voices. 
    \item We aim for it to capture a neutral voice. 
    \item We aim for it to capture no voice. 
    \item We do not aim for it to capture any voice. 
\end{itemize}

\noindent
[SQ36] Does your system or model capture a particular voice? 

\begin{itemize}[noitemsep,topsep=0pt,parsep=0pt,partopsep=0pt]
    \item It captures a non-neutral, specific voice (a particular persona or style).
    \item It captures an assemblage of voices. 
    \item It captures a neutral voice. 
    \item It does not capture any voice. 
    \item It could be capturing a voice but I’m not sure what it is.
\end{itemize}

\noindent
[SQ37] Is neutrality of voice something your system or model should embody? 

\begin{itemize}[noitemsep,topsep=0pt,parsep=0pt,partopsep=0pt]
    \item Definitely yes, yes, maybe, no, definitely no, I don’t know 
\end{itemize}

\subsection{Final thoughts}

\noindent
[SQ38] How do you think NLG evaluation could be improved? {\bf R:} Write-in

\noindent
[SQ39] How would you characterize your experience with practices and metrics aimed at ethical AI? This does not have to be related to your NLG work. 

\begin{itemize}[noitemsep,topsep=0pt,parsep=0pt,partopsep=0pt]
    \item No, I have no experience, I have a bit of experience, I have worked on ethical AI related issues, Ethical AI is the focus of my work 
\end{itemize}

\noindent
[SQ40] If you would like enter the raffle drawing for one of the ten \$50 amazon gift cards, for anonymity purposes, after submitting this form you will be provided with a link to another form to fill in your email address and enter the raffle. For this, please also write down a key phrase here, which you will also be asked to re-enter on the raffle form. We will only be use this key phrase to validate that the raffle participants have completed the survey.  Please don’t use a key phrase that is associated any accounts. {\bf R:} Write-in

\section{Comprehensive Survey Results}
\begin{figure*}[!h]
    \centering
    \includegraphics[width=0.95\textwidth]{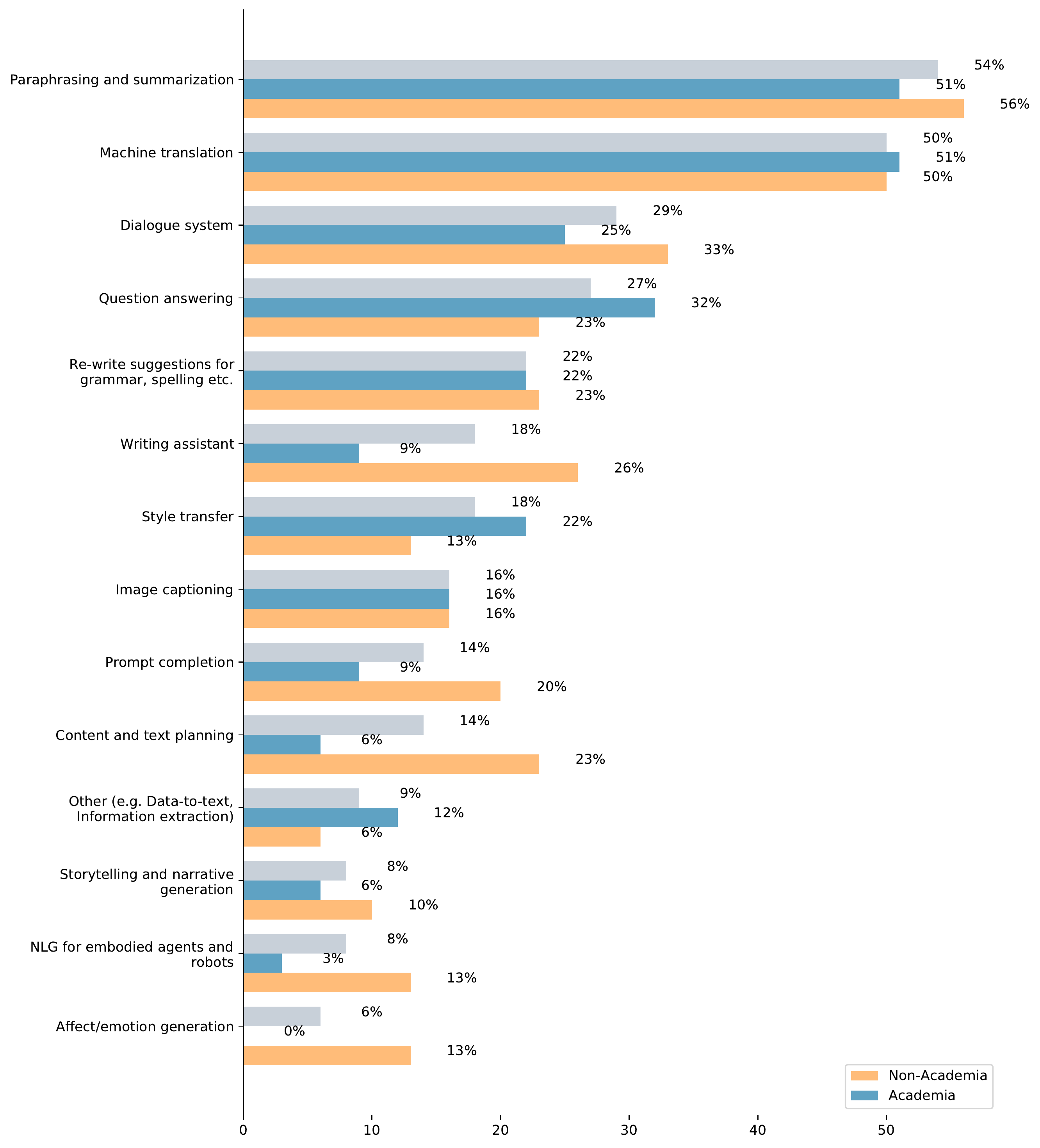} 
    \vspace{-10pt}
    \caption{Question 6: What type of NLG system(s) or task(s) have you worked on or with? Check all that apply.}
    \label{fig:Q10}
\end{figure*}

\begin{figure*}[tbh]
    \centering
    \includegraphics[width=0.95\textwidth]{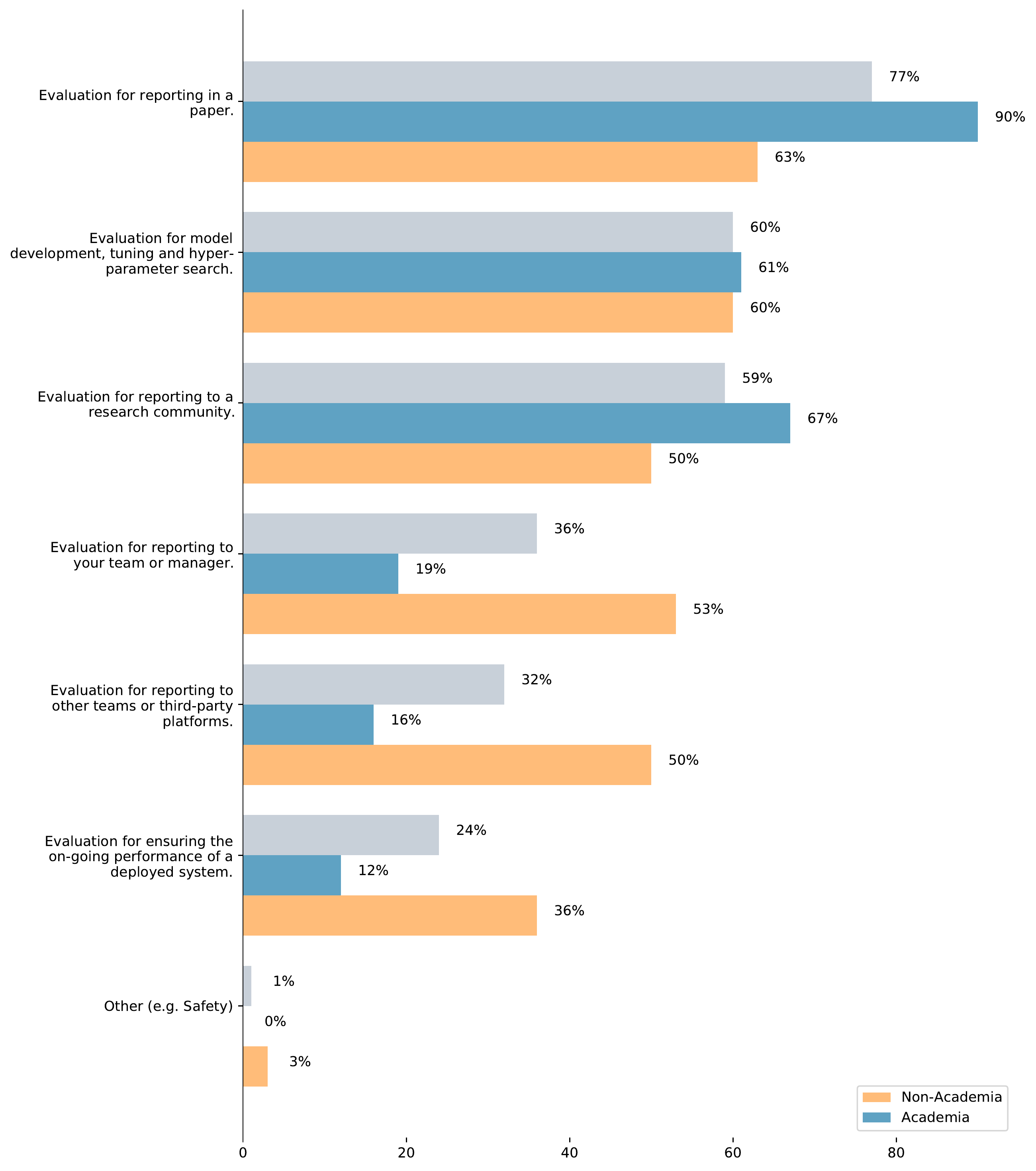}
    \caption{Question 13: When you evaluate your system/model, what are the goals of this evaluation? Check all that apply.}
    \label{fig:Q13}
\end{figure*}

\begin{figure*}[tbh]
    \centering
    \includegraphics[width=\textwidth]{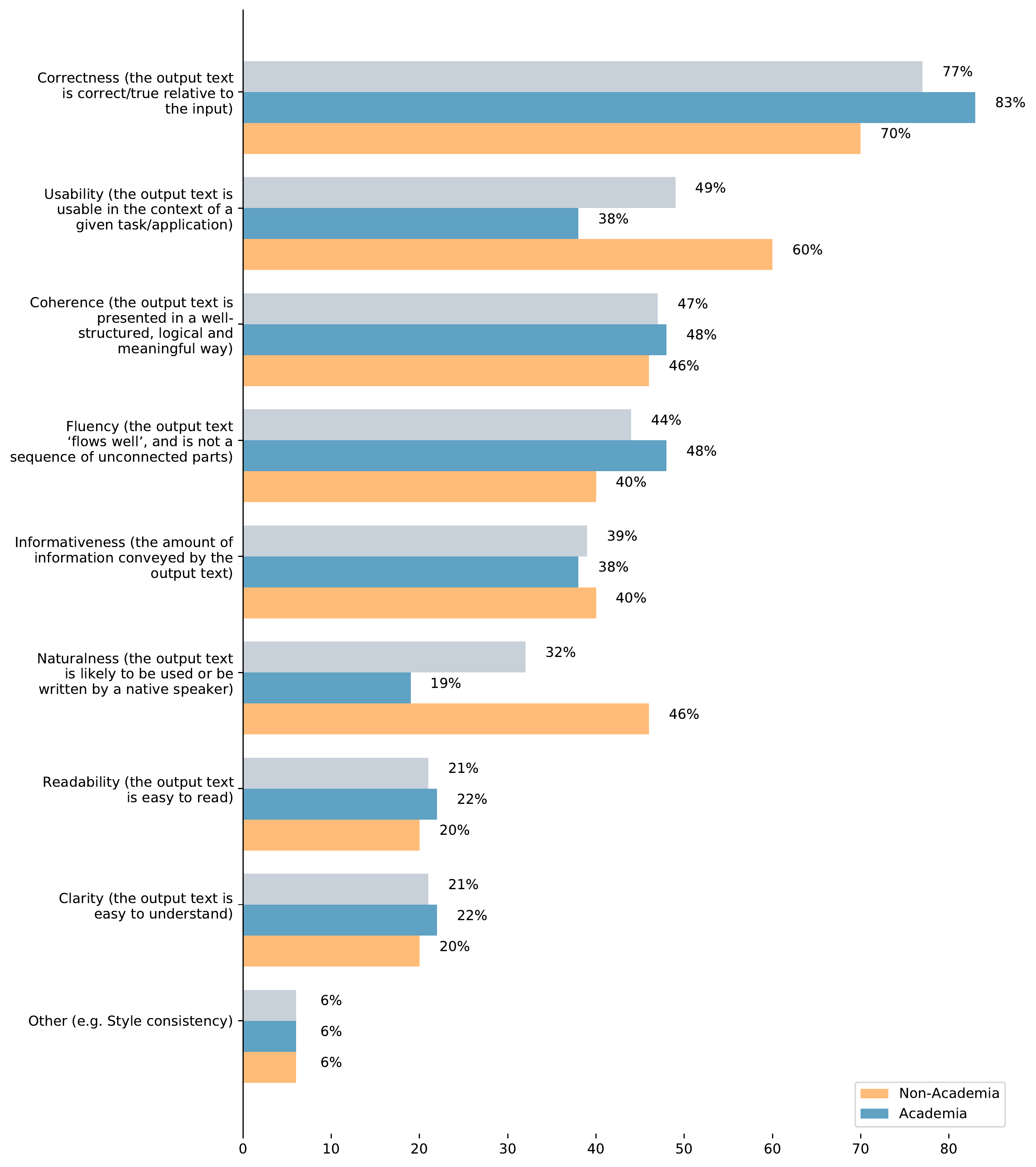}
    \caption{Question 14: Of the following, which are the most important performance criteria to evaluate for your NLG system or task? Select up to three or write your own. For more detailed definitions, please refer to the appendix here: https://aclanthology.org/2020.inlg-1.23.pdf }
    \label{fig:Q14}
\end{figure*}

\begin{figure*}[tbh]
    \centering
    {\includegraphics[width=\textwidth]{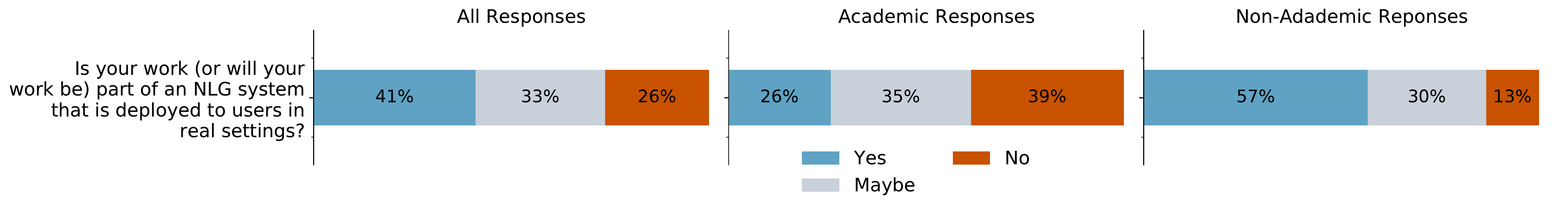} }
    \qquad
    \caption{Question 16: Is your work (or will your work be) part of an NLG system that is deployed to users in real settings (users who use the system for their own reasons, self-incentivized, and are not paid as test subjects/crowd workers)?}
    \label{fig:Q16}
\end{figure*}

\begin{figure*}[tbh]
    \centering
    {\includegraphics[width=\textwidth]{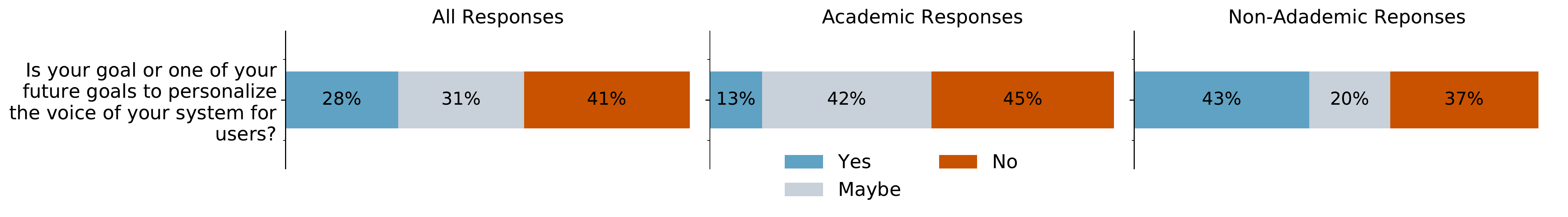} }
    \qquad
    \caption{Question 17: Is your goal or one of your future goals to personalize the voice of your system for users? By voice we mean that it may capture a particular style of speaking or writing, or a particular persona. }%
    \label{fig:q17}%
\end{figure*}

\begin{figure*}[tbh]
    \centering
    {\includegraphics[width=\textwidth]{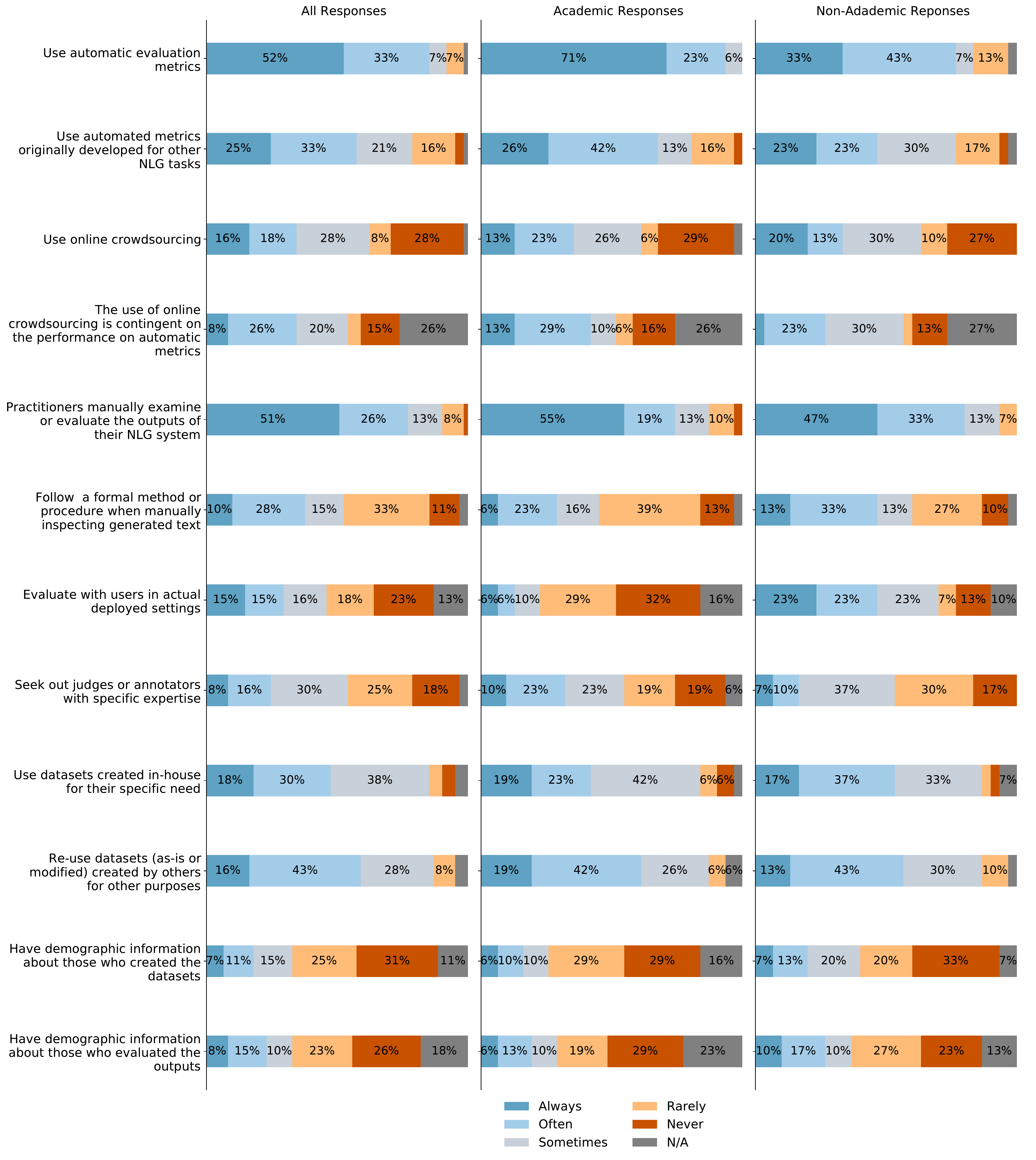} }
    \qquad
    \caption{Question 18: For the following statements, please answer how often you or others on your team do the following.}%
    \label{fig:q18}%
\end{figure*}

\begin{figure*}[tbh]
    \centering
    {\includegraphics[width=\textwidth]{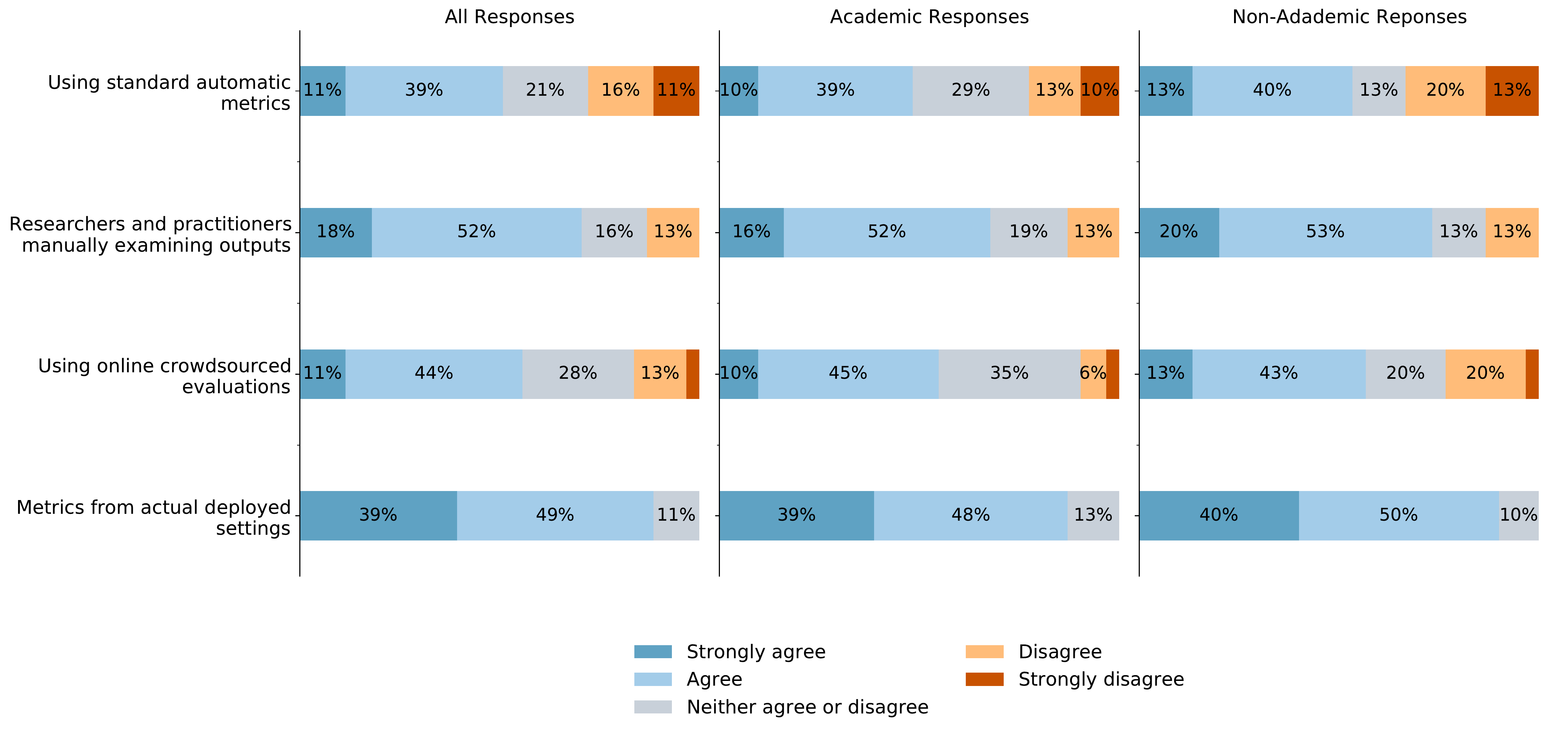} }
    \caption{Question 20: For the system or task, you described earlier, do you agree or disagree that the following evaluation methods are reliable ways to assess performance? }%
    \label{fig:q20}%
\end{figure*}

\begin{figure*}[tbh]
    \centering
    {\includegraphics[width=\textwidth]{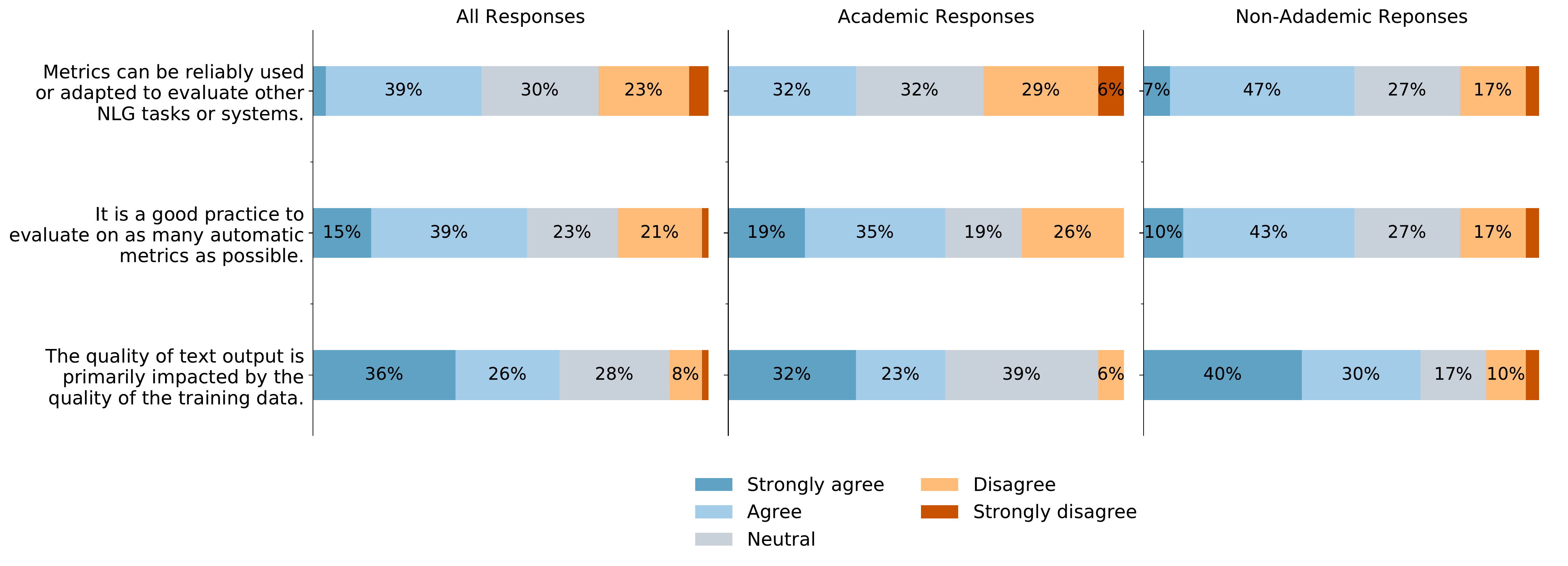} }
    \caption{Question 22: In general, please answer if you agree or disagree with the following statement about evaluating NLG systems or tasks}%
    \label{fig:q22}%
\end{figure*}

\begin{figure*}[tbh]
    \centering
    {\includegraphics[width=\textwidth]{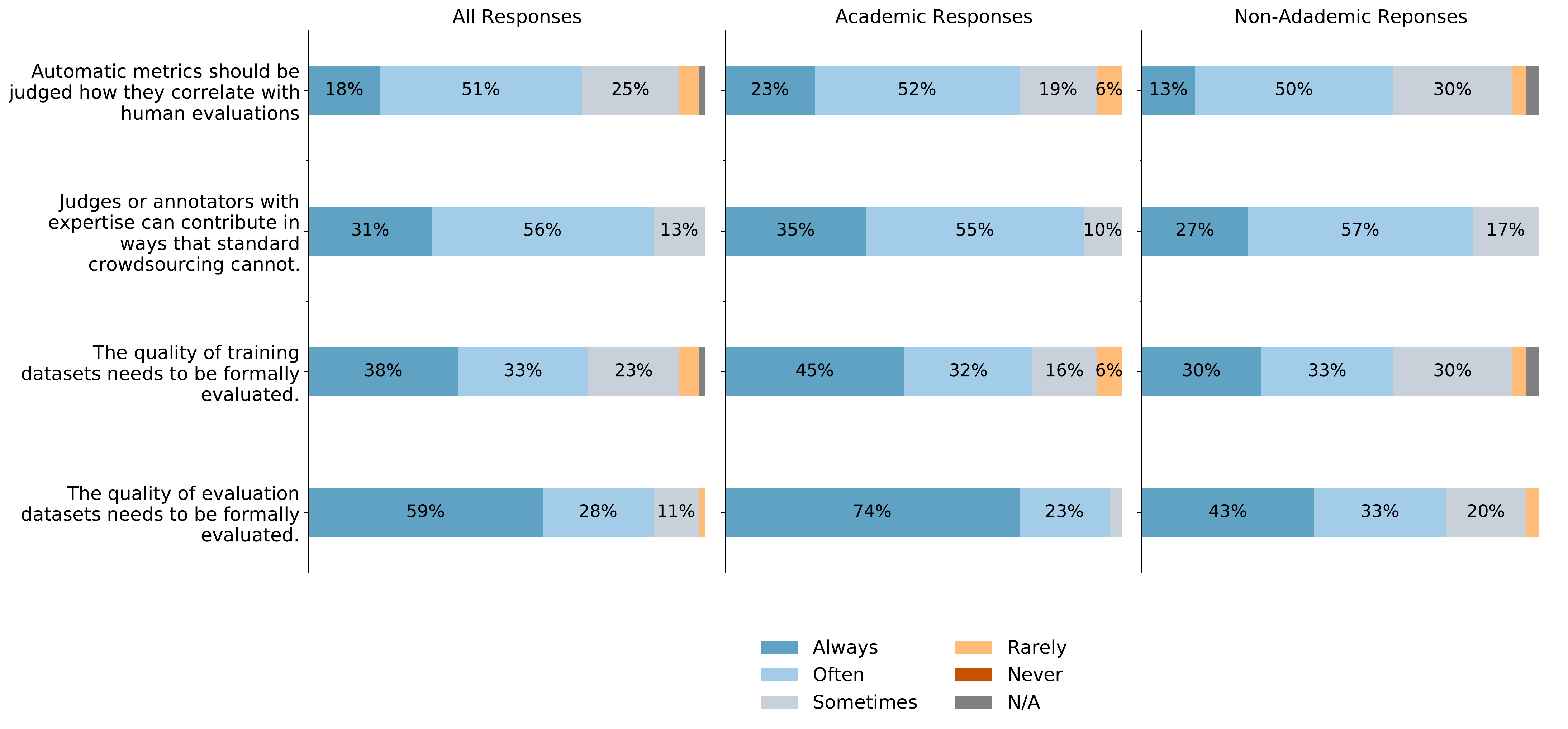} }
    \caption{Question 23: In general, please answer how often the following statements are true:}%
    \label{fig:q23}%
\end{figure*}

\begin{figure*}[tbh]
    \centering
    {\includegraphics[width=\textwidth]{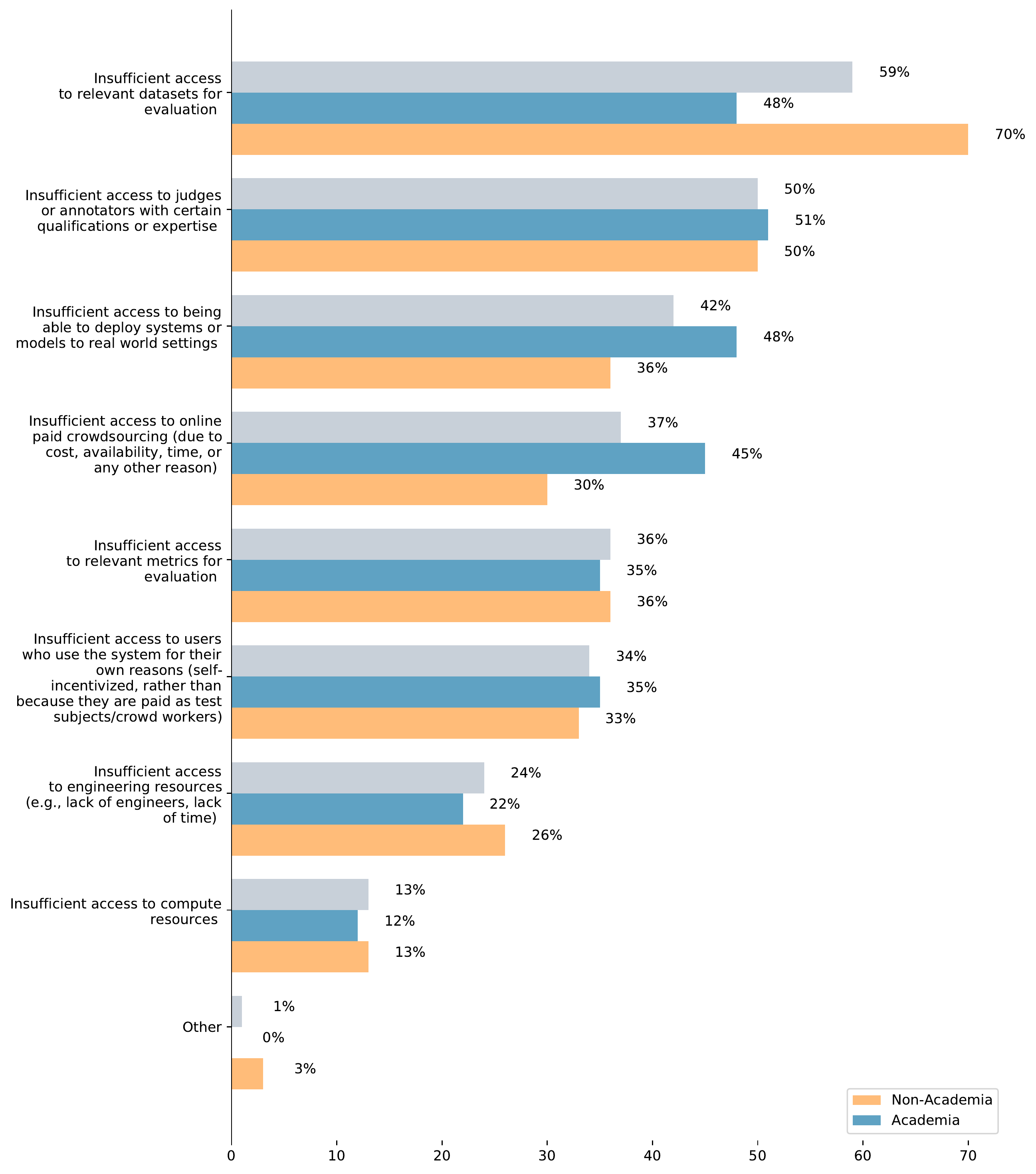} }
    \caption{Question 24: Rank which of the following resources currently limit the evaluation for your particular system or task the most. Top choice being the most limiting resource. Visualizing top 3 choices.}%
    \label{fig:q24}%
\end{figure*}

\begin{figure*}[tbh]
    \centering
    {\includegraphics[width=\textwidth]{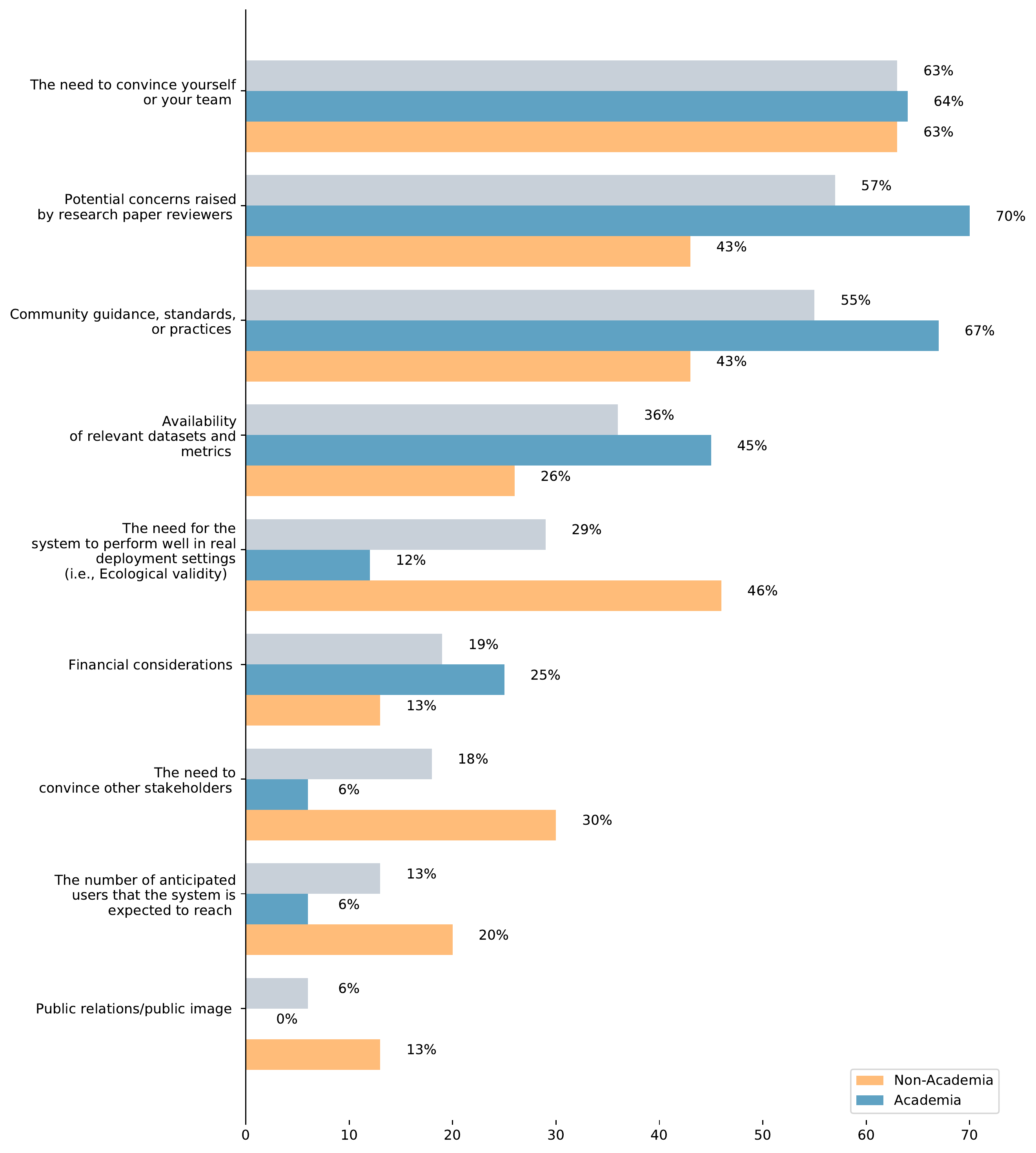} }
    \caption{Question 26: Rank which of the following considerations currently guide your evaluation practices the most. Top choice being the most influential consideration. Visualizing top 3 choices.}%
    \label{fig:q26}%
\end{figure*}

\begin{figure*}[tbh]
    \centering
    {\includegraphics[width=\textwidth]{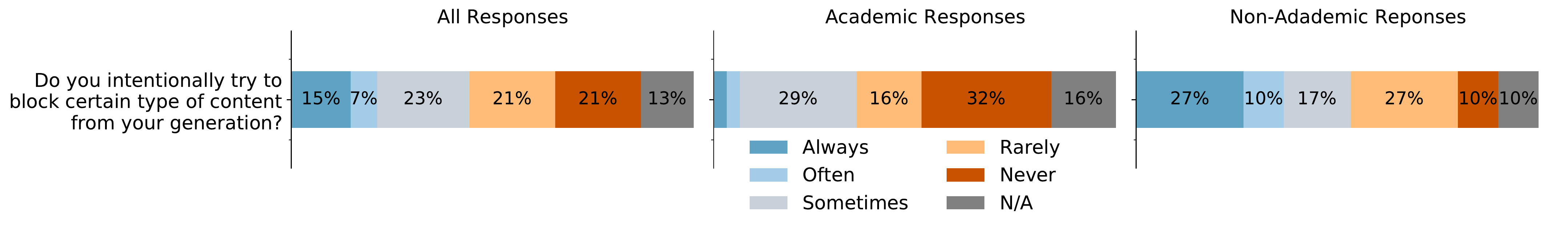} }
    \caption{Question 30: Do you intentionally try to block certain type of content from your generation? (e.g., by using blocklists or classifiers, cleaning the training data, etc.) }%
    \label{fig:q30}%
\end{figure*}

\begin{figure*}[tbh]
    \centering
    {\includegraphics[width=\textwidth]{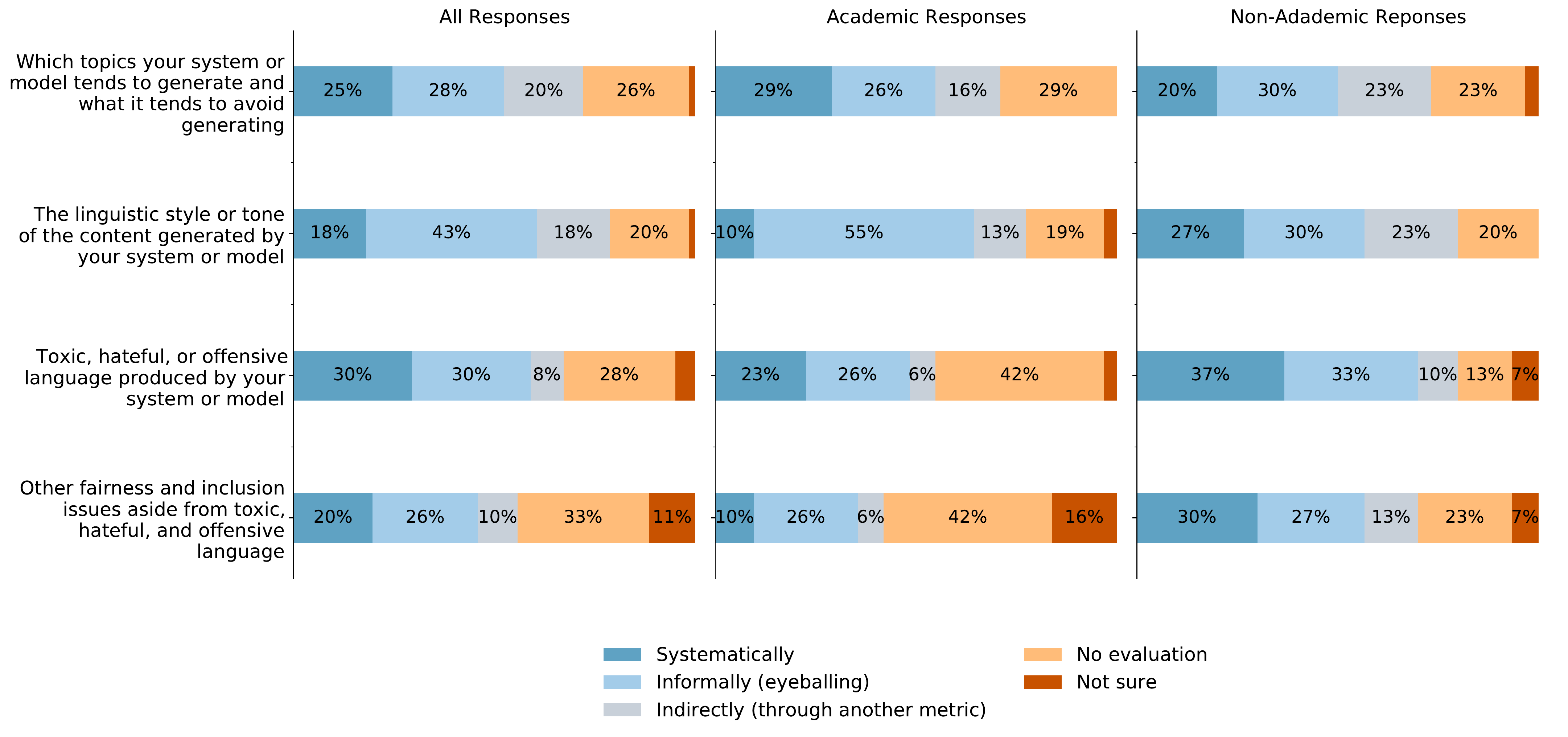} }
    \qquad
    \caption{Question 31: For the following questions, for your system, how do you measure the following? Fairness and Inclusion states that NLG systems should treat all people equally, empowering and engaging everyone by providing equal benefit and access to e.g., opportunities and resources.}%
    \label{fig:q31}%
\end{figure*}

\begin{figure*}[tbh]
    \centering
    {\includegraphics[width=\textwidth]{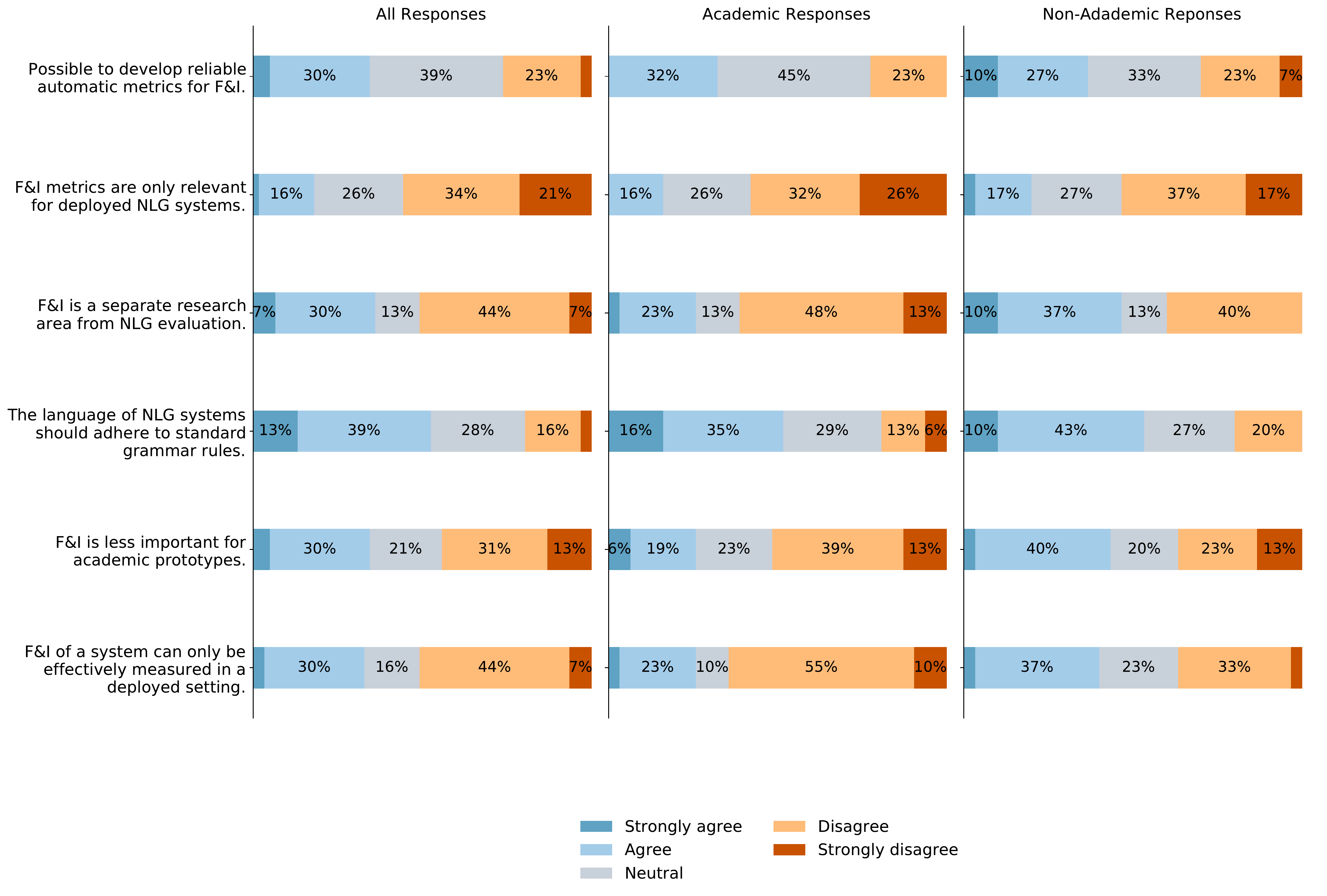} }
    \qquad
    \caption{Question 33: In general, for the following statements, please answer if you agree or disagree with each statement for NLG systems’ evaluation. }%
    \label{fig:q33}%
\end{figure*}

\begin{figure*}[tbh]
    \centering
    {\includegraphics[width=\textwidth]{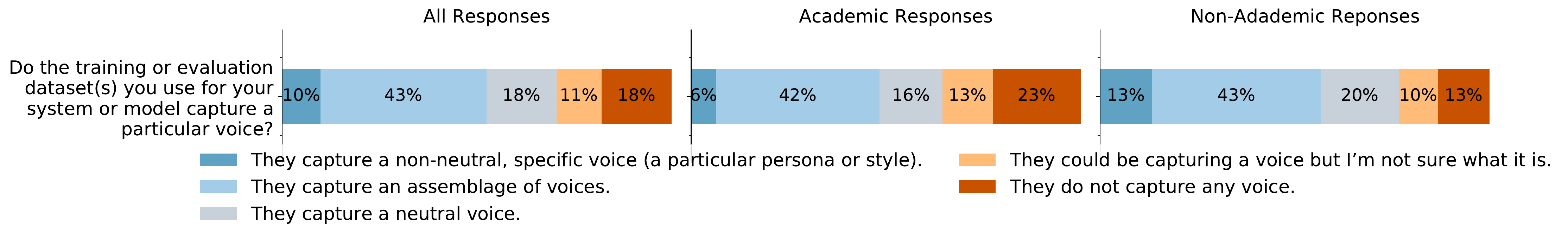} }
    \qquad
    {\includegraphics[width=\textwidth]{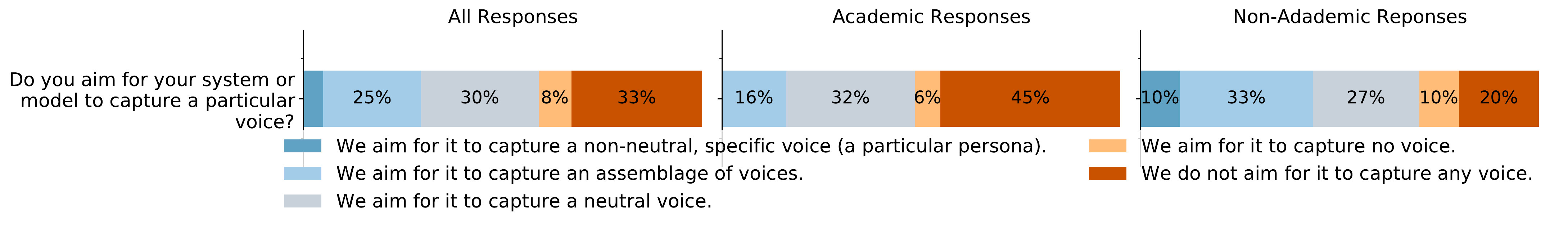} }
    \qquad
    {\includegraphics[width=\textwidth]{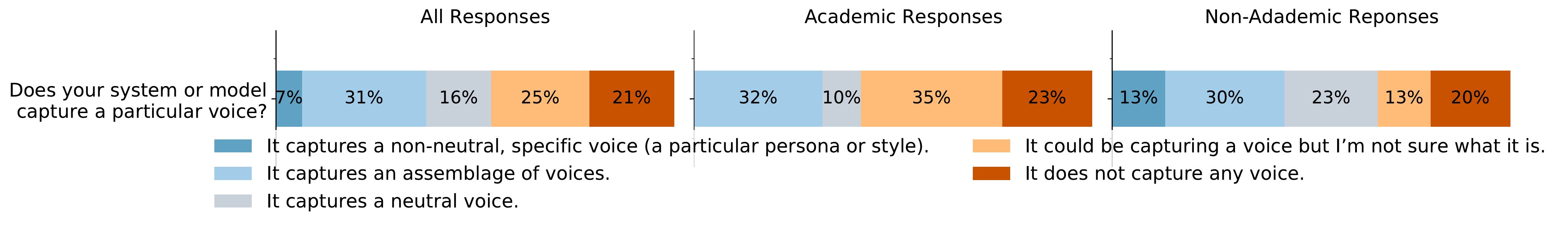} }
    \qquad
    {\includegraphics[width=\textwidth]{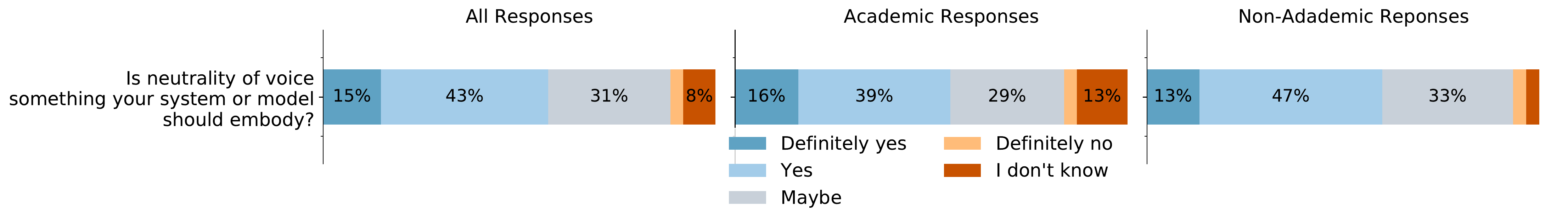} }
    \caption{Questions 34 - 37 on whether a NLG system or model captures a particular voice. By voice we mean that it may capture a particular style of speaking or writing, or a particular persona. }%
    \label{fig:q34}%
\end{figure*}

\begin{figure*}[tbh]
    \centering
    {\includegraphics[width=\textwidth]{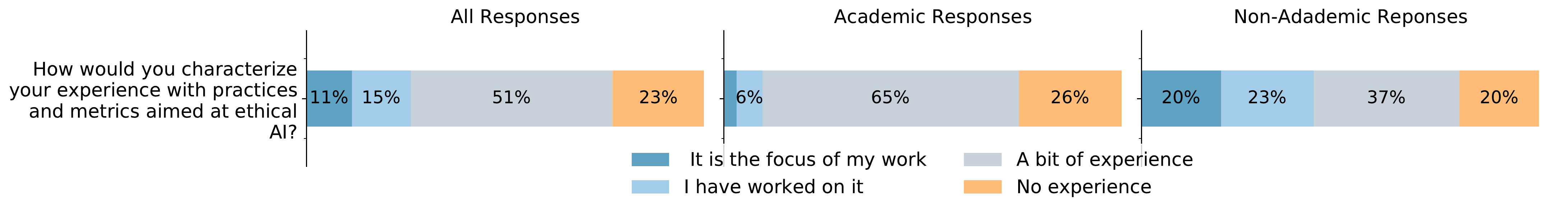}}
    \caption{Question 39: How would you characterize your experience with practices and metrics aimed at   ethical AI? This does not have to be related to your NLG work. }%
    \label{fig:q39}%
\end{figure*}

\end{document}